\pdfoutput=1

\documentclass{article}
\usepackage[accepted]{icml2018}
\usepackage{pdfpages}

\usepackage{microtype}
\usepackage{graphicx}
\usepackage{graphicx, caption, subcaption}
\usepackage{booktabs} 

\usepackage{hyperref}

\usepackage{amsmath}
\usepackage{xfrac}
\usepackage[ruled,vlined,algo2e]{algorithm2e}
\usepackage{amsthm}
\theoremstyle{plain}

\theoremstyle{definition}
\newtheorem{definition}{Definition}
\theoremstyle{plain}
\newtheorem{theorem}{Theorem}
\newtheorem{prop}{Proposition}

\newtheorem{lemma}{Lemma}
\usepackage{bbm}
\usepackage{wrapfig}
\usepackage{xcolor}
\usepackage{enumitem}
\usepackage{array}
\usepackage{multirow}
\usepackage{natbib}
\usepackage{xr}
\newcommand{\specialcell}[2][c]{%
  \begin{tabular}[#1]{@{}c@{}}#2\end{tabular}}

\icmltitlerunning{Multiclass Universum SVM}

\begin{document}

\twocolumn[
\icmltitle{Multiclass Universum SVM}




\begin{icmlauthorlist}
\icmlauthor{Sauptik Dhar}{to}
\icmlauthor{Vladimir Cherkassky}{goo}
\icmlauthor{Mohak Shah}{to,ed}
\end{icmlauthorlist}

\icmlaffiliation{to}{LG Silicon Valley Lab, Santa Clara, CA , USA.}
\icmlaffiliation{goo}{University of Minnesota, MN, USA.}
\icmlaffiliation{ed}{University of Illinois at Chicago, IL, USA}

\icmlcorrespondingauthor{Sauptik Dhar}{sauptik.dhar@gmail.com}

\icmlkeywords{Universum learning, Multiclass SVM, span bound, histogram of projections}

\vskip 0.3in
]



\printAffiliationsAndNotice{}  

\begin{abstract}
We introduce Universum learning for multiclass problems and propose a novel formulation for multiclass universum SVM (MU-SVM). We also propose an analytic span bound for model selection with $\sim 2-4 \times$ faster computation times than standard resampling techniques. We empirically demonstrate the efficacy of the proposed MU-SVM formulation on several real world datasets achieving $>$ 20\% improvement in test accuracies compared to multi-class SVM.

\end{abstract}

\section{Introduction} \label{intro}
Many applications of machine learning involve analysis of sparse high-dimensional data, where the number of input features is larger than the number of data samples. Such settings are typically seen in several real life applications in domains such as, healthcare, autonomous driving, prognostics and  health management etc.~ \cite{cherkassky07}. Such high-dimensional data sets present new challenges for most learning problems. Novel data intensive deep architectures are naturally not suited for such scenarios \cite{goodfellow2016}. Recent studies have shown Universum learning to  be particularly effective for such high-dimensional low sample size data settings~ \cite{sinz08,chen09,dhar15,lu14,qi14,shen12,wang14,zhang08,xu15,xu16,zhu16a,chen2017a,dhar2017}. However, most such studies are limited to binary classification problems. On the other hand, many practical applications involve classification of more than two categories. In order to incorporate \textit{a priori} knowledge (in the form of universum data) for such applications, there is a need to extend universum learning for multiclass problems. 

In this paper we focus on formulating the universum learning for multiclass SVM under balanced settings with equal misclassification costs. Researchers have proposed several methods to solve a multiclass SVM problem. Typically these methods follow two basic approaches ~\cite{hsu02,wang14b}. The first approach follows an \textit{Error Correcting Output Code} (ECOC) based setting ~\cite{dietterich1995}, where several binary classifiers are combined to solve the multiclass problem viz., one-vs-one, one-vs-all, directed acyclic graph SVM \cite{platt99}. Previous works, such as~\cite{sinz07b,chen09} which follow this setting, focus on the binary universum learning paradigm and only provide ``some hints" for their extensions to the multiclass problems. An alternative to the ECOC based setting is the \textit{direct approach}, where the entire multiclass problem is solved through a single larger optimization formulation  \cite{vapnik98,crammer02,weston98}. Recently, \cite{zhang17} adopted such a direct approach for universum learning under a probabilistic framework using a logistic loss function. This paper also adopts such a direct approach, but proposes an alternate universum learning framework that utilizes an SVM like loss function following \cite{crammer02}, and introduces the Multiclass Universum SVM (MU-SVM) formulation. The proposed framework allows for: a) an efficient implementation for MU-SVM using existing multiclass SVM solvers (Section \ref{Compute}), and b) deriving practical analytic error bounds for model selection (Section \ref{ModSel}). Further, compared to ECOC based approaches, we provide a unified framework for multiclass learning under universum settings, with similar (or better) performance accuracies  (see Appendix B.1). 

The main contributions of this paper are as follows: 
\begin{itemize}
\item[1.] We formalize the notion of universum learning for SVM under multiclass settings, and propose a novel \textit{direct} formulation called Multiclass Universum SVM (MU-SVM) (in Section \ref{MUSVM}). The proposed MU-SVM formulation has the neat property that it reduces to: i) standard (C\&S) multiclass SVM in absence of universum data and ii) binary U-SVM formulation \cite{weston06} for two-class problems (Section \ref{MUSVM}, Proposition \ref{prop1}). This consolidates the propriety of MU-SVM as the apt extension for multiclass SVM under universum settings.
\item[2.] The proposed formulation has a desirable structure that renders the MU-SVM formulation solvable through any state-of-art multiclass SVM solvers (Section \ref{Compute}, Proposition \ref{prop2}). 
\item[3.] We provide a new Span definition for multiclass formulations, and derive a leave-one-out bound for MU-SVM (Section \ref{ModSel}, Theorem \ref{theorem1}). Under additional assumptions, we provide a computationally efficient version of the leave-one-out error bound (Section \ref{ModSel}, Theorem \ref{theorem2}), which presents a practical mechanism for model selection.
\item[4.] Empirical results are provided in support of the proposed strategy (Section \ref{results})  
\end{itemize}
Finally, conclusions are presented in Section \ref{conc}. 

Note that, a shorter version of this work is available in \cite{dhar2016universum}. Compared to \cite{dhar2016universum}, this paper includes additional proofs and results as highlighted below,
\begin{itemize}
\item This paper provides the new Propositions \ref{prop1}, \ref{prop2} \& \ref{prop3}.
\item We provide a new leave-one-out bound in Theorem \ref{theorem1} without any assumptions. 
\item Under the assumptions in Section \ref{ModSel}, we provide a stricter leave-one-out error bound, which holds for both \textit{Type} 1 \& 2 support vectors.
\item Exhaustive results for all the claims are provided for additional data sets.   
\end{itemize}




\section{Multiclass SVM} \label{MSVM}
\begin{wrapfigure}{r}{4.2cm}
\includegraphics[width=4.2cm]{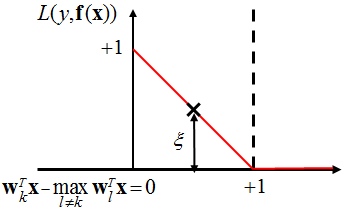}
\caption{Loss function for multiclass SVM with $f_k(\mathbf{x})=\mathbf{w}_k^{\top}\mathbf{x}$. A sample $(\mathbf{x},y=k)$ lying inside the margin is penalized linearly using the slack variable $\xi$.}\label{fig1}
\vspace{-0.1cm}
\end{wrapfigure}  

This section provides a brief description of the multiclass SVM formulation following \cite{crammer02}. Given i.i.d training samples $(\mathbf{x}_i,y_i)_{i=1}^n$, with $\mathbf{x} \in \Re^d$ and $y \in \{1,\ldots, L\}$ ; where $n$ = number of training samples, $d$ = dimensionality of the input space and $L$ = total number of classes, the task of a multiclass classifier is to estimate a vector valued function $\mathbf{f} = [f_1,\ldots,f_L]$ for predicting the class labels for future unseen samples $(\mathbf{x},y)$ using the decision rule $\hat{y} = \underset{l=1,\ldots,L}{\text{argmax}}\; f_l(\mathbf{x})$. The C\&S multiclass SVM is a widely used formulation which generalizes the concept of large margin classifier for  multiclass problems. This multiclass SVM setting employs a special margin-based loss (similar to the hinge loss), $\mathcal{L}(y,\mathbf{f}(\mathbf{x})) = [\underset{l}{max}(f_l(\mathbf{x})+1-\delta_{yl})-f_y(\mathbf{x})]_{+}$ where $[a]_+ = max(0,a)$ and $\delta_{yl} = \left\{
\begin{array}{l l}
    1;\quad  y=l \\
    0;\quad  y\neq l
\end{array}\right.$ (see Fig \ref{fig1}). Here, for any sample $(\mathbf{x},y = k)$, having $\mathcal{L}(y,\mathbf{f}(\mathbf{x})) = 0$ ensures a margin-distance of `+1' for the correct prediction i.e. $f_k(\mathbf{x})-f_l(\mathbf{x}) \geq 1 ; \forall l \neq k$. The SVM multiclass formulation (for linear parameterization) is provided below: \vspace{-0.3cm}
\begin{align} \label{eq1SVM}
\underset{\mathbf{w}_1 \ldots \mathbf{w}_L ,\boldsymbol\xi}{\text{min}} & \quad \frac{1}{2} \sum \limits_{l=1}^L\|\mathbf{w}_l \|_{2}^{2} \quad+\quad C\sum\limits_{i=1}^n \xi_{i} &&\\
s.t. & \quad (\mathbf{w}_{y_{i}}-\mathbf{w}_l)^\top \mathbf{x}_i \geq e_{il} - \xi_{i} ;\quad e_{il} = 1-\delta_{il} &&\nonumber\\ 
&\quad i=1 \ldots n ,\quad l = 1 \ldots L  && \nonumber 
\end{align} 
here,  $f_l(\mathbf{x}) = \mathbf{w}_l^{\top}\mathbf{x} $ . Note that training samples falling inside the margin border (`+1') are linearly penalized using the slack variables $\xi_i \geq 0, \; i=1 \ldots n$ (as shown in Fig \ref{fig1}). These slack variables contribute to the empirical risk for the multiclass SVM formulation $R_{emp}(\mathbf{w})=\sum \limits_{i=1}^{n} \xi_i$. The SVM \footnote{We refer to the C \& S formulation in \eqref{eq1SVM} as SVM throughout.} formulation attempts to strike a balance between minimization of the empirical risk and the regularization term. This is controlled through the user-defined parameter $C \geq 0$.

\section{Multiclass Universum SVM} 
\subsection{Multiclass U-SVM formulation} \label{MUSVM}

\begin{wrapfigure}{r}{4.5cm}
\includegraphics[width=4.5cm]{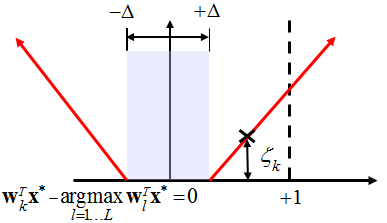}
\caption{Loss function for universum samples $\mathbf{x}^{*}$ for $k^{th}$ class decision boundary $\mathbf{w}_k^{\top}\mathbf{x}^{*} - \underset{l=1 \ldots L}{\text{max}} \mathbf{w}_l^{\top}\mathbf{x}^{*} = 0$. A sample lying outside the $\Delta$- insensitive zone is penalized linearly using the slack variable $\zeta_k$.}\label{fig2}
\end{wrapfigure}

The idea of Universum learning was introduced by \cite{vapnik98,vapnik06} to incorporate a priori knowledge about admissible data samples. The Universum learning was introduced for binary classification, where in addition to labeled training data we are also given a set of unlabeled examples from the Universum. The Universum contains data that belongs to the same application domain as the training data. However, these samples are known not to belong to either class. In fact, this idea can also be extended to multiclass problems. For multiclass problems in addition to the labeled training data we are also given a set of unlabeled examples from the Universum. These Universum samples are known not to belong to \textit{any} of the classes in the training data. For example, if the goal of learning is to discriminate between handwritten digits 0, 1, 2,...,9; one can introduce additional `knowledge' in the form of handwritten letters A, B, C, ... ,Z. These examples from the Universum contain certain information about handwriting styles, but they cannot be assigned to any of the classes (0 to 9). Also note that, Universum samples do not have the same distribution as labeled training samples. These unlabeled Universum samples are introduced into the learning as contradictions and hence should lie close to the decision boundaries of all the classes ${1 \ldots L}$. This argument follows from \cite{vapnik06,weston06}, where the universum samples lying close to the decision boundaries are more likely to falsify the classifier. To ensure this, we incorporate a $\Delta$ - insensitive loss function for the universum samples (shown in Fig \ref{fig2}). This $\Delta$ - insensitive loss forces the universum samples to lie close to the decision boundaries (`0' in Fig. \ref{fig2}). Note that, this idea of using a $\Delta$ - insensitive loss for Universum samples has been previously introduced in \cite{weston06} for binary classification. However, different from \cite{weston06}, here the $\Delta$ - insensitive loss is introduced for the decision boundary for all the classes i.e. $\mathbf{w}_k^{\top} \mathbf{x}^{*} - \underset{l=1 \ldots L}{\text{max}}\mathbf{w}_l^{\top} \mathbf{x}^{*} = 0 \; ; \forall k = 1 \ldots L$.  This reasoning motivates the new multiclass Universum-SVM (MU-SVM) formulation where:
\begin{itemize}
\item[--] Standard hinge loss is used for the training samples (shown in Fig. \ref{fig1}). This loss forces the training samples to lie outside the `+1' margin border.
\item[--] The universum samples are penalized by a $\Delta$ - insensitive loss (see Fig. \ref{fig2}) for the decision functions of all the classes $\mathbf{f} = [f_1,\ldots,f_L]$.
\end{itemize}
This leads to the MU-SVM formulation. Given training samples $\mathcal{T}:=(\mathbf{x}_i,y_i)_{i=1}^n$, where $y_i \in \lbrace 1,\ldots,L\rbrace$ and additional unlabeled universum samples $\mathcal{U}:=(\mathbf{x}_{i^\prime}^{*})_{i^\prime=1}^m$. Solve\footnote{Throughout this paper, we use index $i, j$ for training samples, $i^\prime$ for universum samples and $k, l$ for the class labels.},
\begin{flalign}\label{eq2USVM}
&\underset{\mathbf{w}_1 \ldots \mathbf{w}_L ,\boldsymbol\xi,\boldsymbol\zeta}{\text{min}}  \quad \frac{1}{2} \sum \limits_{l=1}^L\|\mathbf{w}_l \|_{2}^{2} + C\sum\limits_{i=1}^n \xi_{i} + C^{*}\sum\limits_{i^\prime =1}^m \sum\limits_{k=1}^L \zeta_{i^\prime k}&& \nonumber \\
&s.t. \quad \forall  i=1\ldots n, \quad i^\prime = 1 \ldots m &&\\
& \quad \quad (\mathbf{w}_{y_{i}}-\mathbf{w}_l)^\top \mathbf{x}_i \geq e_{il} - \xi_{i} ; \;  e_{il} = 1-\delta_{il}, \; l=1 \ldots L && \nonumber \\
& \quad \quad |(\mathbf{w}_k^\top \mathbf{x}_{i^\prime}^{*}-\underset{l=1 \ldots L}{\text{max}}\mathbf{w}_l^\top \mathbf{x}_{i^\prime}^{*})| \leq \Delta + \zeta_{i^\prime k}; \quad k = 1 \ldots L  && \nonumber \\
& \quad \quad \zeta_{i^\prime k} \geq 0 ,\quad \delta_{il} = \left\{
\begin{array}{l l}
    1;\quad  y_i=l \\
    0;\quad  y_i\neq l
\end{array}\right. &&\nonumber
\end{flalign}
Here, for the $k^{th}$ class decision boundary the universum samples $(\mathbf{x}_{i^\prime}^{*})_{i^\prime =1}^m$ that lie outside the $\Delta$ - insensitive zone are linearly penalized using the slack variables $\zeta_{i^\prime k} \geq 0, \; i^\prime = 1\ldots m$. The user-defined parameters $C,C^{*} \geq 0$ control the trade-off between the margin size, the error on training samples, and the contradictions (samples lying outside $\pm\Delta$ zone) on the universum samples. Note that for $C^{*} = 0$ eq. \eqref{eq2USVM} reduces to the multiclass SVM classifier.

\begin{prop} \label{prop1}
For binary classification  L = 2, \eqref{eq2USVM} reduces to the standard U-SVM formulation in \cite{weston06} with $\mathbf{w} = \mathbf{w}_1 -\mathbf{w}_2$ and $b = 0$. 
\end{prop}

\subsection{Computational Implementation of MU-SVM} \label{Compute}

This section describes computational implementation of the MU-SVM formulation \eqref{eq2USVM}. Here, for every universum sample $\mathbf{x}_{i^\prime}^{*}$  we create $L$ artificial samples belonging to all the classes, i.e. $(\mathbf{x}_{i^\prime}^{*},y_{i^\prime 1}^{*}=1),\ldots, (\mathbf{x}_{i^\prime}^{*},y_{i^\prime L}^{*}=L)$ as below,
\begin{flalign} \label{eq3USVMTrans}
(\mathbf{x}_i,y_i) &= \left\{
\begin{array}{l l l} 
(\mathbf{x}_i,y_i) \quad & i=1 \ldots n \\
(\mathbf{x}_{i^\prime}^{*},y_{i^\prime l}^{*}) \quad & i=n+1 \ldots n+mL;\\
& i^\prime =1 \ldots m; \quad l=1 \ldots L 
\end{array} \right. && \nonumber \\
e_{il} &= \left\{
\begin{array}{l l l} 
e_{il} \quad & i=1 \ldots n; \quad l=1 \ldots L  \\
-\Delta(1-\delta_{i^\prime l}) \quad & i=n+1 \ldots n+mL; \\
& i^\prime =1 \ldots m; \quad l=1 \ldots L
\end{array} \right. &&  \\
C_i &= \left\{
\begin{array}{l l l} 
C \quad & i=1 \ldots n \\
C^{*} \quad & i=n+1 \ldots n+mL; \\
& i^\prime =1 \ldots m ; \; l=1 \ldots L
\end{array} \right. && \nonumber 
\end{flalign}   
\begin{prop} \label{prop2}
Under transformation \eqref{eq3USVMTrans}, the MU-SVM formulation in eq. \eqref{eq2USVM} can be exactly solved using,
\begin{align}\label{eq4USVM2SVM}
\underset{\mathbf{w}_1 \ldots \mathbf{w}_L ,\mathbf{\xi}}{\text{min}} & \quad \quad \frac{1}{2} \sum \limits_{l=1}^L\|\mathbf{w}_l \|_{2}^{2} \quad+\quad \sum\limits_{i=1}^{n+mL} C_i \ \xi_{i} &&\\
s.t. & \quad (\mathbf{w}_{y_{i}}-\mathbf{w}_l)^\top \mathbf{x}_i \geq e_{il} - \xi_{i}&& \nonumber \\ 
& \quad i=1 \ldots n+mL ,\quad l = 1 \ldots L && \nonumber
\end{align}
\end{prop}
The formulation~\eqref{eq4USVM2SVM} has the same form as~\eqref{eq1SVM} except that the former has additional $mL$ constraints for the universum samples. Like most other SVM solvers, the MU-SVM formulation in~\eqref{eq4USVM2SVM} is also solved in its dual form as shown in Algorithm~\ref{alg1} see \cite{hsu02}. Hence, the computational complexity is same as solving a multiclass SVM formulation (in~\eqref{eq1SVM}) with $n+mL$ samples. Most off-the-shelf multiclass SVM solvers can be used for solving the proposed MU-SVM.

\begin{algorithm} \SetAlgoNoLine 
1. Given training $(\mathbf{x}_i,y_i)_{i=1}^n$ and universum $(\mathbf{x}_{i^\prime}^{*})_{j=1}^m$ \\
2. Transform \eqref{eq3USVMTrans} and solve \eqref{eq5USVKer}, \
\begin{align}\label{eq5USVKer}
\underset{\boldsymbol\alpha}{\text{max}} & \quad W(\boldsymbol \alpha)= - \frac{1}{2} \sum \limits_{i,j} \sum \limits_{l} \alpha_{il} \alpha_{jl} K(\mathbf{x}_i,\mathbf{x}_j) -  \sum\limits_{i,l}\alpha_{il}e_{il} && \nonumber \\
s.t. & \quad \sum \limits_{l} \alpha_{il} =0 && \\ 
& \quad \alpha_{i,l} \leq C_i \quad  \text{if} \quad l=y_i ; \quad \alpha_{i,l} \leq 0 \quad  \text{if} \quad l \neq y_i  && \nonumber \\
& \quad i, j = 1\ldots n+mL, \; l= 1 \ldots L \nonumber &&
\end{align} \
3. Obtain the class label using the following decision rule:\quad $\hat{y} = \underset{l}{\text{argmax}}\sum \limits_{i}\alpha_{il}K(\mathbf{x}_i,\mathbf{x})$\
\caption{MU-SVM (dual form) \label{alg1}}
\end{algorithm}

\subsection{Model Selection} \label{ModSel}
As presented in~\eqref{eq5USVKer}, the current MU-SVM algorithm has four tunable parameters: $C, C^* ,\text{kernel parameter, and }  \Delta$. So in practice, multiclass SVM may yield better results than MU-SVM, simply because it has an inherently simpler model selection. Successful application of the proposed MU-SVM heavily depends on the optimal tuning of its model parameters. This paper adopts a simplified strategy for model selection previously used in \cite{cherkassky11}. This mainly involves two steps,
\begin{itemize}
\item[a.]  First, perform optimal tuning of the $C$ and kernel parameters for multiclass SVM classifier. This step equivalently performs model selection for the parameters specific only to the training samples in the MU-SVM formulation \eqref{eq2USVM}.
\item[b.] Second, tune the parameter $\Delta$ while keeping $C$ and kernel parameters fixed (as selected in Step a). Parameter $C^*/C = \frac{n}{mL}$ is kept fixed throughout the paper to ensure equal contribution of training and universum samples in the optimization formulation.
\end{itemize}
This strategy selects an MU-SVM solution (in step b) close to a given SVM solution (selected in step a). The model parameters in Steps (a) \& (b) are typically selected through resampling techniques such as, leave one out (l.o.o) or stratified cross-validation approaches \cite{shah11}. Of these approaches, l.o.o provides an \textit{almost} unbiased estimate of the test error \cite{luntz1969,scholkopf2001learning}. However, on the downside it is very computationally intensive. In this paper, we propose a new analytic bound for the leave-one-out error for MU-SVM formulation. The proposed bound can be used for model selection in Steps (a) \& (b) and provides a computational edge over standard resampling techniques. Detailed discussion regarding this new l.o.o error bound is provided next.

Note that, the l.o.o formulation with the $t^{th}$ training sample dropped is the same as in \eqref{eq5USVKer} with an additional constraint $\alpha_{tl} = 0 ; \quad  \forall l$. Then, the l.o.o error is given as: $ R_{l.o.o} = \frac{1}{n}\sum \limits_{t=1}^{n} \mathbbm{1}[y_t \neq \hat{y}_t] $, where $\hat{y}_t=  \underset{l}{\text{arg max}}\sum \limits_{i}\alpha_{il}^t K(\mathbf{x}_i,\mathbf{x}_t)$ is the predicted class label for the $t^{th}$ sample and $\boldsymbol\alpha^t = [\underset{\boldsymbol\alpha_1^t}{\underbrace{\alpha_{11}^t,\ldots,\alpha_{1L}^t}},\ldots,\underset{\boldsymbol\alpha_t^t = \mathbf{0}}{\underbrace{\alpha_{t1}^t = 0,\ldots,\alpha_{tL}^t = 0}},\ldots]$ is the l.o.o solution. In this paper, we follow a strategy very similar to the one used in \cite{vapnik00}, and derive the new l.o.o bound for the MU-SVM formulation in \eqref{eq5USVKer}. The necessary prerequisites are presented next.
\begin{definition}(Support vector categories)\label{def1} 
\begin{itemize}[nosep]
\item[1.] A support vector obtained from eq.~\eqref{eq5USVKer} is called a \textit{Type 1} support vector if $0 < \alpha_{iy_i} < C_i$. This is represented as, $SV_1 = \{\ i \ | 0< \alpha_{iy_i} < C_i \}$ 
\item[2.] A support vector obtained from eq.~\eqref{eq5USVKer} is called a \textit{Type 2} support vector if $
\alpha_{iy_i}=C_i$. This is represented as, $SV_2 = \{\ i \ | \alpha_{iy_i} = C_i \}$ 
\end{itemize}
\end{definition} 
The set of all support vectors are represented as, $SV = SV_1 \cup SV_2$. Similarly, the set of support vectors for \textit{l.o.o} solution is given as $SV^t$. Under definition~\eqref{def1} we have,

\begin{lemma} \label{lemma1}
If in leave-one-out procedure a Type 1 support vector $\mathbf{x}_t$ is classified incorrectly, then we have,
\begin{align}
S_t \ max(\sqrt{2}D,\frac{1}{\sqrt{C}})\geq 1 \nonumber
\end{align}
where, \begin{flalign} \label{spandef}
S_t^2 = &\quad \underset{\boldsymbol\beta}{\text{min}}\quad  \sum \limits_{i,j}(\sum \limits_{l}\beta_{il}\beta_{jl})K(\mathbf{x}_i,\mathbf{x}_j) &&\\
s.t. & \quad \alpha_{il} - \beta_{il} \leq C_i; \quad \forall \lbrace (i\neq t,l) |\ 0<\alpha_{il} < C_i;\ l = y_i \rbrace &&\nonumber\\
& \quad \alpha_{il} - \beta_{il} \leq 0; \quad \forall\  \lbrace (i\neq t,l) |\ \alpha_{il} < 0;\ l \neq y_i \rbrace &&\nonumber\\
&\quad \beta_{il} =0 ; \quad \forall i \notin SV_1-\{t\} \; \forall l=1 \ldots L && \nonumber\\
&\quad \beta_{tl} =\alpha_{tl} ;\quad \forall l=1 \ldots L && \nonumber\\
&\quad \sum \limits_{l} \beta_{il} =0 &&\nonumber
\end{flalign}
$S_t$ := Span of the Type 1 support vector $\mathbf{x}_t$ \\
$D$ := Diameter of the smallest hypersphere containing all training samples.
\end{lemma}
This leads to the following upper bound on the l.o.o error.
\begin{theorem} \label{theorem1}
The leave-one-out error is upper bounded as: 
\begin{flalign} \label{eqloo_noassume}
R_{l.o.o} \leq & \quad \frac{1}{n} ( |\Psi_1 | + |\Psi_2 |)  && \\
\Psi_2 := & \Big\{ t \in SV_1 \cap \mathcal{T} \; | S_t \ max(\sqrt{2}D,\frac{1}{\sqrt{C}})\geq 1 \Big\} && \nonumber \\
\Psi_1 := & \Big\{ \;t \in  SV_2 \cap \mathcal{T} \Big\} ;\quad |\cdot | := \text{Cardinality of a set} && \nonumber
\end{flalign} 
and  $\mathcal{T}:=$ Training Set.
\end{theorem}

Following Theorem 1, it is desirable to select a model with a) lower number of Type 2 training support vectors and, b) smaller span for the type 1 training support vectors. Roughly, for a fixed number of type 2 support vectors a solution with smaller span value (for the type 1 training support vectors) could yield lower test error. The following proposition shows how the universum samples influence these span values in \eqref{spandef}.   

\begin{prop} \label{prop3}
If the Type 1 training support vectors i.e. $ t \in SV_1 \cap \mathcal{T} $ for SVM and MU-SVM solutions remain same, then $ \; S_t^{SVM} \geq S_t^{MU-SVM} \; ; \forall t  \in SV_1 \cap \mathcal{T} $. 
\end{prop}

Loosely speaking, for cases where the type of training support vectors remain same, introducing universum samples through the MU-SVM formulation could result in smaller span values and better generalization for future test data compared to standard SVM solution.  

Now, Theorem \ref{theorem1} provides an analytic tool for model selection with small l.o.o error. Here, the right hand side of \eqref{eqloo_noassume} serves as a leave-one-out error estimate, and the goal is to select a model parameter which minimizes this value.  However, the practical utility of \eqref{eqloo_noassume} is limited due to the significant computational complexity involved in estimating the span of the type 1 training support vectors  $\sim O(n+mL)^4$ (worst case). Next, we provide a more computationally attractive alternative to the above l.o.o bound. 

\textbf{Assumption}: For the MU-SVM solution,
\begin{itemize}[nosep] 
\item[i] The set of support vectors of the \textit{Type1} and \textit{Type2} categories remain the same during the leave-one-out procedure.
\item[ii] The dual variables of the \textit{Type1} support vectors have only two active elements i.e. $\forall \boldsymbol\alpha_i ~\textit{s.t.}~ \lbrace 0<\alpha_{iy_i}<C_i \rbrace \ \exists \ k \neq y_i ~~\textit{s.t.} ~~ \alpha_{ik} = - \alpha_{iy_i}$.
\end{itemize}




\begin{lemma} \label{lemma2}
Under the above assumptions the following equality holds for both Type 1\& 2 support vectors,
\begin{align}\label{eq8span_assume}
S_t^2 =& [\boldsymbol\alpha_t^{\top} \sum\limits_{i \in SV} \sum\limits_{l} \alpha_{il} K(\mathbf{x}_i,\mathbf{x}_t) &&\\
& \quad  \quad \quad \quad - \alpha_{ty_t}\mathbf{g}_{y_tk}^{\top} \sum \limits_{i \in SV^t} \sum \limits_{l} \alpha_{il}^t K(\mathbf{x}_i,\mathbf{x}_t)] && \nonumber
\end{align}  with, \quad  $S_t^2 =\{\underset{\boldsymbol\beta}{\text{min}}\ \sum \limits_{i,j}(\sum \limits_{l}\beta_{il}\beta_{jl})K(\mathbf{x}_i,\mathbf{x}_j)|\ \boldsymbol\beta_{t} =\boldsymbol\alpha_{t} ;\ \sum \limits_{l} \beta_{il} =0\ ; (i,j)\in SV_1\}$\quad and \quad  $\mathbf{g}_{y_t k} =[0,\ldots \underset{y_t}{1},\ldots,\underset{k^{th}}{-1},\ldots,0]; \; k = \underset{q \neq y_t}{argmax} \sum\limits_{j} \alpha_{jq}^t K(\mathbf{x}_j,\mathbf{x}_t)$ 
\end{lemma}

Now $S_t$ can be efficiently computed using lemma \eqref{lemma3}. 
\newline
\begin{lemma} \label{lemma3}
Under Assumptions (i) \& (ii)
\begin{flalign} 
S_t^2 &= \left\{
\begin{array}{l l l} 
\boldsymbol\alpha_t^{\top} [(\mathbf{H}^{-1})_{\mathbf{tt}}]^{-1} \boldsymbol\alpha_t  & t \in SV_1 \cap \mathcal{T} \\
\boldsymbol\alpha_t^{\top} [K(\mathbf{x}_t,\mathbf{x}_t)\otimes \mathbf{I}_{L}-\mathbf{K}_t^{T}\mathbf{H}^{-1}\mathbf{K}_t] \boldsymbol\alpha_t  & t \in SV_2 \cap \mathcal{T}
\end{array} \right. && \nonumber 
\end{flalign}
\begin{flalign}
\text{here,} \quad \mathbf{H} & := \begin{bmatrix}
    \mathbf{K}_{SV_1} \otimes \mathbf{I}_{L}       & \mathbf{A}^{\top} \\
    \mathbf{A}     & \mathbf{0}
\end{bmatrix}; \nonumber &&\\
\mathbf{A} & := \mathbf{I}_{|SV_1|} \otimes (\mathbf{1}_L)^{\top} ; \quad \quad  \mathbf{1}_L =[\underset{L\ elements}{\underbrace{1 \ 1 \ldots \ 1}}] \nonumber && \\
\quad (\mathbf{H}^{-1})_{\mathbf{tt}} & := \text{sub-matrix of } \mathbf{H}^{-1} \text{for indices } &&\nonumber\\
& \quad \quad i = (t-1)L+1  \ldots tL &&\nonumber \\
\quad \mathbf{K}_{SV_1} \quad & := \text{Kernel matrix of Type 1 support vectors}. &&\nonumber \\
\mathbf{K}_t \quad &= [(\mathbf{k}_t^T \otimes \mathbf{1}_L) \; \; \mathbf{0}_{L\times |SV_1|}]^T && \nonumber
\end{flalign}
where, $\mathbf{k}_t = n_{|SV_1| \times 1}$ dim vector where ith element is $K(\mathbf{x}_i,\mathbf{x}_t), \forall \mathbf{x}_i \in SV_1$ ; and $\otimes$ is the Kronecker product.
\end{lemma}  
Finally, we have,
\begin{theorem} \label{theorem2}
Under the Assumptions (i) \& (ii) the leave-one-out error is upper bounded as: 
\begin{flalign} \label{eq9loo_assume}
R_{l.o.o} \leq & \; \frac{1}{n}|\Psi_3 | && \\
\Psi_3 = & \Big\{ t \in SV \cap \mathcal{T} \; | \; S_t^2 \geq \boldsymbol\alpha_t^{\top} \sum\limits_{i \in SV} \sum\limits_{l} \alpha_{il} K(\mathbf{x}_i,\mathbf{x}_t)\Big\}\bigg] \nonumber &&
\end{flalign}
and  $\mathcal{T}:=$ Training Set ; and $S_t:=$ defined in Lemma \ref{lemma3}
\end{theorem}  

Note that, similar to \cite{vapnik00}, the assumptions (i) \& (ii) are not satisfied in most cases. Nevertheless, Theorem 2 provides a good approximation of the l.o.o procedure (see Section \ref{mselexp}). In addition, compared to Theorem \ref{theorem1}, it provides the following advantages, 
\begin{itemize}
\item[--] Eq. \eqref{eq9loo_assume} is valid for both type 1 \& 2 training support vectors and typically results in a stricter bound.
\item[--] Span computation for all support vectors requires inverting the $\mathbf{H}$ - matrix only once (Lemma \ref{lemma3}). This results to an overall cost of $O(n+mL)^3$ for computing \eqref{eq9loo_assume} and provides a computation edge over \eqref{eqloo_noassume} which involves a cost of $O(n+mL)^4$.
\end{itemize} 
Empirical results for model selection using Theorem \ref{theorem2} are provided in Section \ref{mselexp}.

\section{Empirical Results} \label{results}


\begin{table}
\centering
\caption{Real-life datasets.} \label{data}
\begin{small}
\begin{sc}
\begin{tabular}{ccc}  
\hline
\abovespace\belowspace
Dataset & Train/Test size & Dimension \\ 
\hline
\abovespace\belowspace 
 GTSRB & \specialcell{300 / 1500\\(100 / 500 per class)} &\specialcell{1568\\(HOG Features)}\\ 
 \abovespace\belowspace
 ABCDETC & \specialcell{600 / 400\\(150 / 100 per class)} & \specialcell{10000\\(100 x 100 Pixel)}\\ 
\abovespace\belowspace
 ISOLET & \specialcell{500 / 500\\(100 / 100 per class)} & 617 \\ \hline
\end{tabular}
\end{sc}
\end{small}
\end{table}

\subsection{Datasets and Experimental settings} \label{dataexp}
Our empirical results use three real life datasets : \\ 
\textit{German Traffic Sign Recognition Benchmark} (GTSRB) \cite{stall12}:  The goal here is to identify the traffic signs for the speed-zones `30',`70' and `80'. Here, the images are represented by their histogram of gradient (HOG 1) features. The experimental setting is provided in Table \ref{data}. For this data we use three kinds of Universum: 
\begin{itemize}
\item[--] \textit{Random Averaging}: synthetically created by first selecting a random traffic sign from each class (`30',`70' and `80') in the training set and averaging them. 
\item[--] \textit{Non-Speed} : all other \textit{non}-speed zone traffic signs.
\item[--] \textit{Sign `priority-road'}: An exhaustive search over several non-speed zone traffic signs showed this universum to provide the best performance (Appendix B.3) 
\end{itemize} 

\textit{Handwritten characters} (ABCDETC) \cite{weston06}: The data consists images of handwritten digits `0-9', uppercase `A-Z', lowercase letters `a-z' and some additional symbols: ! \; ? \; , \; . \; ; \; : \; = \; - \; + \; / \; / \; ( \; ) \; \$ \; \% \; " \; @. The goal here is to identify the handwritten digits `0' - `3' based on their pixel values. We use four different types of universum: \textit{Upper}: `A - Z' , \textit{Lower}: `a - z' , \textit{Symbols}: all additional symbols and \textit{Random Averaging} (RA) obtained by randomly averaging the training samples. 

\textit{Speech-based Isolated Letter Recognition} (ISOLET) \cite{fanty1991}: This is a speech recognition dataset where 150 subjects spoke the name of each letter `a - z' twice. The goal is to identify the spoken letters `a' - `e' using the spectral coefficients, contour features, sonorant features, presonorant features, and post-sonorant features. We use two different types of universum: \textit{Others}, which consists of all other speech i.e. `f' -`z' and \textit{Random Averaging} (RA). 

Note that, to simplify our analysis (in Section \ref{svmvsusvm}) we used a subset of the training classes. However, similar results can be expected using all the training classes (Appendix B.2). Our initial experiments suggest that  linear parameterization is optimal for the GTSRB dataset; hence only linear kernel has been used for it. For the ABCDETC and ISOLET datasets an RBF kernel of the form $K(\mathbf{x}_i,\mathbf{x}_j) = exp(-\gamma \Vert \mathbf{x}_i - \mathbf{x}_j \Vert^2)$ with $\gamma = 2^{-7}$ provided optimal results for SVM. For all the experiments model selection is done over the range of parameters, $C = [10^{-4},\ldots,10^{3}]$ , $C^{*}/C = \frac{n}{mL}$ and $\Delta =[0,0.01,0.05,0.1]$ using stratified 5-Fold cross validation. 

\subsection{Results} 
\subsubsection{Comparison between SVM vs. MU-SVM} \label{svmvsusvm}


Performance comparisons between SVM and MU-SVM for the different types of Universum are shown in Table \ref{tabResults}.  The table shows the average test error over 10 random training/test partitioning of the data in similar proportions as shown in Table \ref{data}. As seen from Table \ref{tabResults}, MU-SVM provides better generalization than SVM. In fact, for certain universum types, like \textit{Priority-Road} for GTSRB, \textit{RA} for ABCDETC and ISOLET; MU-SVM significantly outperforms the multiclass SVM model. In such cases, the performance gains improve significantly upto $\sim 20-25 \%$ with the increase in number of universum samples, and stagnates for a significantly large universum set size. This indicates that for sufficiently large universum set size the effectiveness of MU-SVM depends mostly on the type (statistical characteristics) of the universum data. For a better understanding of such statistical characteristics, we adopt the technique of `\textit{histogram of projections}' originally introduced for binary classification in \cite{cherk10}. However, different from binary classification, here we project a training sample $(\mathbf{x},y = k)$ onto the decision space for that class i.e. $\mathbf{w}_k^{\top}\mathbf{x}-\underset{l \neq k}{\text{max}}\ \mathbf{w}_l^{\top}\mathbf{x}=0$ and the universum samples onto the decision spaces of all the classes. Finally, we generate the histograms of the projection values for our analysis. Further, in addition to the histograms, we also generate the frequency plot of the predicted labels for the universum samples.

\begin{table}
\centering
\caption{Mean ($\pm$ standard deviation) of the test errors (in \%) over 10 runs of the experimental setting in Table \ref{data}.} 
\label{tabResults}
\tabcolsep=0.10cm
\begin{small}
\begin{sc}
\begin{tabular}{ccccc}  
\hline
\abovespace\belowspace 
\multirow{2}{*}{GTSRB}&\multicolumn{4}{c}{No. of Universum samples}
\\ \cline{2-5}
\abovespace
& MU-SVM & 200 & 500 & 1000 \\ 
\hline
\abovespace
\multirow{3}{*}{\rotatebox{90}{\specialcell{SVM \\ $7.54 \pm 0.82$}}} & \specialcell{Priority\\Road} & $6.97 \pm 1.06$ & $5.52 \pm 0.68$ & $5.51 \pm 0.78$ \\ 
\abovespace
& RA & $7.08\pm 0.71$ & $6.98 \pm 0.93$ & $7.08 \pm 0.43$ \\ 
 \abovespace
& \specialcell{Non-\\Speed} & $7.53 \pm 0.64$ & $7.46 \pm 0.6$ & $7.3 \pm 0.93$ \\ \hline
\abovespace\belowspace 
ABCDETC &  & 200 & 500 & 1000 
\\ \cline{2-5}
\hline
\abovespace
\multirow{5}{*}{\rotatebox{90}{\specialcell{SVM \\ $27.1 \pm 3.5$}}} & \specialcell{Upper} & $26.5 \pm 3.9$ & $ 26.1 \pm 3.6$ & $26.1 \pm 4.0$ \\ 
\abovespace
& \specialcell{Lower} & $25 \pm 3.2$ & $24.2 \pm 3.4$ & $24.2 \pm 3.1$ \\ 
\abovespace
& Symbols & $23.5 \pm 4.3$ & $23.1 \pm 3.2$ & $23.3 \pm 3.2$ \\ 
\abovespace
& RA & $23.2 \pm 4.8$ & $22.2 \pm 3.5$ & $22.1 \pm 3.2$ \\ 
\hline
\abovespace\belowspace
ISOLET &  & 200 & 500 & 1000 
\\ \cline{2-5}
\hline
\abovespace\belowspace
\multirow{3}{*}{\rotatebox{90}{\specialcell{SVM \\ $3.6 \pm 0.3$}}}
& RA & $3.05\pm 0.34$ & $ 2.78 \pm 0.24$ & $2.77 \pm 0.28$ \\ 
\abovespace\belowspace
& Others & $3.50 \pm 0.3$ & $3.31 \pm 0.27$ & $3.31 \pm 0.3$ \\ \hline
\end{tabular}
\end{sc}
\end{small}
\end{table}

\begin{figure*}[h]
\centering
\includegraphics[height=2.4cm, width=17cm]{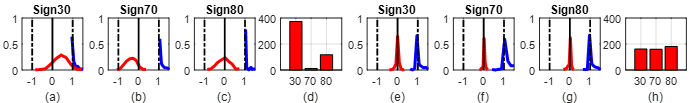}
\caption{Typical histogram of projection for training samples ($n = 300$) (shown in \textcolor{blue}{blue}) and universum samples `\textit{priority-road}' ($m=500$) (shown in \textcolor{red}{red}). SVM decision functions (with $C = 1$) for (a) sign `30'. (b) sign `70'.(c) sign `80'. (d) frequency plot of predicted labels for universum samples using SVM model. MU-SVM decision functions (with $C^*/C = 0.2, \Delta = 0.01$) for (e) sign `30'. (f) sign `70'.(g) sign `80'. (h) frequency plot of predicted labels for universum samples using MU-SVM model.} \label{histprior} 
\end{figure*}

\begin{figure*}
\centering
\includegraphics[height=2.4cm, width=17cm]{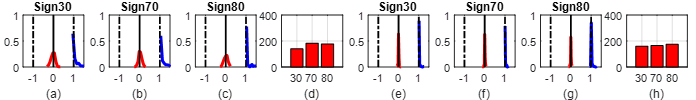}
\caption{Typical histogram of projection for training samples ($n = 300$) (shown in \textcolor{blue}{blue}) and universum samples `\textit{Random Averaging}' ($m=500$) (shown in \textcolor{red}{red}). SVM decision functions (with $C = 1$) for (a) sign `30'. (b) sign `70'.(c) sign `80'. (d) frequency plot of predicted labels for universum samples using SVM model. MU-SVM decision functions (with $C^*/C = 0.2, \Delta = 0$) for (e) sign `30'. (f) sign `70'.(g) sign `80'. (h) frequency plot of predicted labels for universum samples using MU-SVM model.} \label{histra} \vspace{0.3cm}
\includegraphics[height=2.4cm, width=17cm]{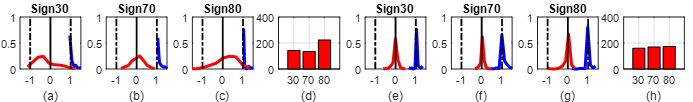}
\caption{Typical histogram of projection for training samples ($n = 300$) (shown in \textcolor{blue}{blue}) and universum samples `\textit{Others}' ($m=500$) (shown in \textcolor{red}{red}). SVM decision functions (with $C = 1$) for (a) sign `30'. (b) sign `70'.(c) sign `80'. (d) frequency plot of predicted labels for universum samples using SVM model. MU-SVM decision functions (with $C^{*}/C = 0.2,\Delta = 0.05$) for (e) sign `30'. (f) sign `70'.(g) sign `80'. (h) frequency plot of predicted labels for universum samples using MU-SVM model.} \label{histothers}
\end{figure*}

Figs. \ref{histprior} shows the typical histograms and frequency plots for the SVM and MU-SVM models for the GTSRB dataset using the `\textit{priority-road}' sign (as universum). As seen from Fig. \ref{histprior}, the optimal SVM model has high separability for the training samples i.e., most of the training samples lie outside the margin borders. In fact, similar to binary SVM \cite{cherk10}, we see data-piling effects for the training samples near the `+1' - margin borders of the decision functions for all the classes. This is typically seen under high-dimensional low sample size settings. However, the universum samples (`\textit{priority-road}') are widely spread about the margin-borders. Moreover, here the universum samples are biased towards the positive side of the decision boundary of the sign `30' (Fig. \ref{histprior}(a)) and hence predominantly gets classified as sign `30'(Fig.\ref{histprior}(d)). As seen from Figs  \ref{histprior}. (e)-(h), applying the MU-SVM model preserves the separability of the training samples and additionally reduces the spread of the universum samples. Such a model exhibits uncertainty on the universum samples' class membership, and uniformly assigns them over all the classes i.e. signs `30',`70' and `80' (Fig. \ref{histprior}(h)). This shows that, the resulting MU-SVM model has higher contradiction (uncertainty) on the universum samples and hence provides better generalization compared to SVM.

Fig \ref{histra} shows the histograms and the frequency plots for SVM and MU-SVM models for RA universum. As shown in Fig \ref{histra} (a), the SVM model already results in a narrow distribution of the universum samples and in turn provides \textit{near} random prediction on the universum samples (Fig. \ref{histra}(d)). Applying MU-SVM for this case provides no significant change to the multiclass SVM solution and hence no additional improvement in generalization (see Table \ref{tabResults} and Fig.\ref{histra} (e)-(h)). 

Finally, we provide the histograms and the frequency plots for SVM and MU-SVM models for the \textit{Non-Speed} Universum samples. In this case, although the universum samples are widely spread about the SVM margin-borders (Figs \ref{histothers}(a)-(c)), yet the uncertainity on the universum samples' class membership is uniform across all the classes (Fig \ref{histothers}(d)). Applying MU-SVM reduces the spread of the universum samples (Figs. \ref{histothers}(e) - (g)). However, it does not significantly increase the contradiction (uncertainity) on the universum samples (compare Figs. \ref{histothers} (d) vs. (h)). Hence, applying MU-SVM does not provide any significant improvement over the SVM model (see Table \ref{tabResults}). The histograms for the other datasets provide similar insights and have been provided in Appendix B.4. 

This section shows that for high-dimensional low sample size settings, MU-SVM provides better generalization than multiclass SVM. Under such settings the training data exhibits large data-piling effects near the margin border (`+1'). For such ill-posed settings, introducing the Universum can provide improved generalization over the multiclass SVM solution. However, the effectiveness of the MU-SVM also depends on the properties of the universum data. Such statistical characteristics of the training and universum samples for the effectiveness of MU-SVM can be conveniently captured using the `histogram-of-projections' method introduced in this paper.

\subsubsection{Effectiveness using Analytic Bound in Theorem \ref{theorem2}} \label{mselexp}

\begin{table}
\centering
\caption{Performance comparisons for model selection using cross validation vs. analytic bound in Theorem \ref{theorem2}. Train/Test partitioning follows Table \ref{data}. No. of universum samples ($m = 1000$). Model parameters used $C^*/C = \frac{n}{mL}, \; \Delta = [0, 0.01, 0.05, 0.1]$}. 
\label{tabBound}
\tabcolsep=0.05cm
\begin{small}
\begin{sc}
\begin{tabular}{cccccc}  
\hline
\abovespace\belowspace 
& &\multicolumn{2}{c}{5-Fold CV} & \multicolumn{2}{c}{Theorem \ref{theorem2}} 
\\ \cmidrule(lr){3-4} \cmidrule(lr){5-6}
\abovespace
& MUsvm & \specialcell{Test Error\\(in $\%$)} & \specialcell{Time\\ ($\times 10^4 sec $)} & \specialcell{Test Error\\(in $\%$)} & \specialcell{Time\\($\times 10^4 sec$)} \\ 
\hline
\abovespace
\multirow{3}{*}{\rotatebox{90}{GTSRB \quad}}&\specialcell{Priority\\Road} & $5.5 \pm 0.6$ & $3.5 \pm 0.3$ & $5.2 \pm 0.4$ & $0.9 \pm 0.1$  \\ 
\abovespace
&RA & $6.9 \pm 0.9$ & $3.7 \pm 0.5$ & $6.9 \pm 0.9$ & $0.8 \pm 0.3$  \\
\abovespace
&\specialcell{Non-\\Speed} & $6.9 \pm 0.9$ & $3.7 \pm 0.7$ & $6.8 \pm 0.8$ & $1.0 \pm 0.5$  \\ 
\hline
\abovespace
\multirow{3}{*}{\rotatebox{90}{ABCDETC \;}}&\specialcell{Upper} & $26.1 \pm 4.0$ & $2.8 \pm 0.1$ & $26.1 \pm 3.7$ & $1.1 \pm 0.1$  \\ 
\abovespace
&Lower & $24.2 \pm 3.1$ & $2.8 \pm 0.1$ & $24.4 \pm 3.2$ & $1.3 \pm 0.1$  \\
\abovespace
&Symbols & $23.3 \pm 3.2$ & $2.6 \pm 0.2$ & $24.1 \pm 3.8$ & $0.9 \pm 0.09$  \\ 
\abovespace
&RA & $22.1 \pm 3.2$ & $2.6 \pm 0.1$ & $22.0 \pm 2.8$ & $0.9 \pm 0.1$  \\
\hline
\abovespace\belowspace
\multirow{2}{*}{\rotatebox{90}{\; ISOLET}}&\specialcell{RA} & $2.8 \pm 0.3$ & $3.1 \pm 0.6$ & $2.6 \pm 0.3$ & $1.9 \pm 0.7$  \\ 
\abovespace\belowspace
&Others & $3.3 \pm 0.3$ & $4.8 \pm 0.9$ & $3.3 \pm 0.3$ & $2.1 \pm 0.5$  \\
\hline
\end{tabular}
\end{sc}
\end{small}
\end{table} 

Next we illustrate the practical utility of the bound in Theorem \ref{theorem2} for model selection. 

\begin{figure}
    \begin{subfigure}[t]{4cm}
      \includegraphics[width=4cm]{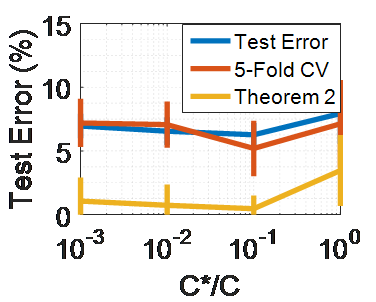}
      \caption{ }
    \end{subfigure}
    \begin{subfigure}[t]{4cm}
      \includegraphics[width=4cm]{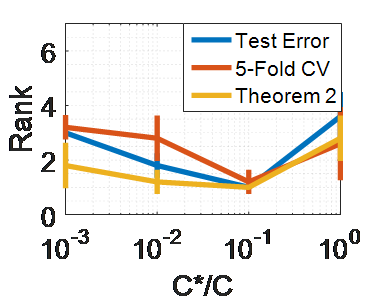}
      \caption{ }
    \end{subfigure}
    \caption{Performance of MU-SVM with \textit{priority-road} universum for the GTSRB dataset. Here, no. of training samples ($ n = 300 $), no. of universum samples ($m = 1000$) (a) Error estimates for the model parameters $C^*/C = [10^{-3},10^{-2},10^{-1},10^0]$, $C = 1, \; \Delta = 0$. (b) Ranking of the model parameters based on the error estimate values over each experiments.} \label{figmselC}
\end{figure}

First, we provide a comparison between the error estimates using 5-Fold cross validation (CV) vs. Theorem \ref{theorem2}\footnote{Note that, Theorem \ref{theorem2} approximates Theorem \ref{theorem1} to provide an upper bound on the l.o.o error. Hence, a good comparison would be between Theorem \ref{theorem2} vs. Theorem \ref{theorem1} and \textit{l.o.o} error. However, results using \textit{l.o.o} and Theorem \ref{theorem1} were prohibitively slow and hence could not be reported in this paper. As an alternative, we compare the error estimates from Theorem \ref{theorem2} with 5-Fold cross validation (CV) and test error. The objective is to illustrate that similar to 5-Fold CV, using Theorem \ref{theorem2} we can obtain the optimal model parameters providing smallest test error.}. For illustration we use the GTSRB dataset under the experimental setting provided in Table \ref{data}. Fig. \ref{figmselC} (a) shows the average error estimates using 5-Fold CV and Theorem \ref{theorem2} as well as the true test error for the MU-SVM model using \textit{priority-road} over the range of parameters $C^*/C = [10^{-3}, 10^{-2}, 10^{-1}, 10^{0}]$ with fixed $\Delta = 0$. The results are obtained over 10 random partitioning of the training/test dataset. Fig. \ref{figmselC} (a) shows that the error estimates using Theorem \ref{theorem2} follows a very similar pattern as 5-Fold CV and test error. This shows that the model parameter $C^{*}/C= 10^{-1}$ that minimizes the l.o.o error estimate in Theorem \ref{theorem2}, also minimizes the test error and 5 Fold CV. Hence, Theorem \ref{theorem2} provides a practical alternative to model selection using resampling techniques. 

\begin{figure}
    \begin{subfigure}[t]{4cm}
      \includegraphics[width=4cm]{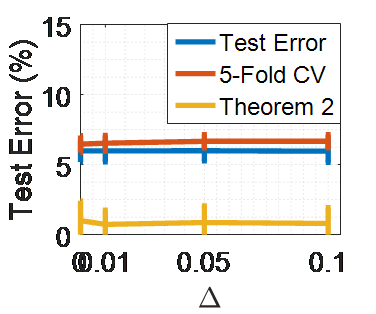}
      \caption{ }
    \end{subfigure}
    \begin{subfigure}[t]{4cm}
      \includegraphics[width=4cm]{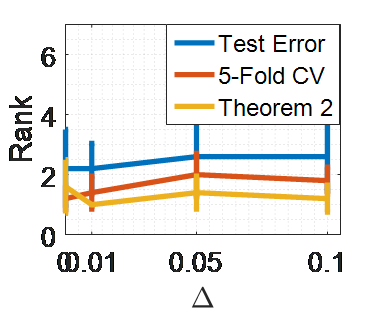}
      \caption{ }
    \end{subfigure}
    \caption{Performance of MU-SVM with \textit{priority-road} universum for the GTSRB dataset. Here, no. of training samples ($ n = 300 $), no. of universum samples ($m = 1000$) (a) Error estimates for the model parameters $\Delta = [0,0.01,0.05,0.1]$, $C =1, C^*/C = \frac{n}{mL} = 0.1$ (b) Ranking of the model parameters based on the error estimate values over each experiments.} \label{figmselG}
\end{figure}

Throughout our results we observe that the error estimates using Theorem \ref{theorem2} are uniformly lower than the 5-Fold CV and test error. This can be attributed to two main reasons. First, for high-dimensional low sample size settings, majority of the training samples lie outside the margin borders (see Figs. \ref{histprior}-\ref{histothers}). This results in a significantly low proportion of training SVs, and hence low l.o.o error in general. Secondly, Theorem \ref{theorem2} holds under additional assumptions (i) \& (ii), and is further constrained compared to Theorem \ref{theorem1}. Hence, Theorem \ref{theorem2} is an under estimator of the loo bound in Theorem \ref{theorem1}. Of course, for the purpose of model selection we are only interested in the pattern, rather than the scale of the error estimates. Hence, such a difference in scale will not impact the model selection. However, to further simplify our illustrations, we also provide a scale invariant ranking curve of the model parameters in Fig. \ref{figmselC}(b). The figure shows the average rankings of the model parameters based on the error estimate values over each experiments. Here, for each experiment we rank the model parameter with the smallest error estimate as $\sim 1$, and the parameter with the largest estimate as $\sim 4$, and average these rank values over the 10 experiments. The parameter with the smallest rank value $\sim 1$ (in Fig. \ref{figmselC}(b)) is typically selected through the model selection strategy. Finally, as seen from Figs. \ref{figmselC} (a) - (b), although different in scale, the error estimates using Theorem \ref{theorem2} correctly captures the pattern of the test error and selects the model parameter with the smallest test error (i.e. $C^*/C = 10^{-1}$).  A similar comparison over the range of parameters $\Delta = [0.001, 0.01, 0.1, 1]$ with fixed $C^*/C = \frac{n}{mL} = 0.3$ is also provided in Fig. \ref{figmselG}. Here, compared to 5 - Fold CV , Theorem \ref{theorem2} correctly selects the optimal parameter $\Delta = 0.01$ with the smallest test error (Fig. \ref{figmselG} (b)).


As seen from Figs. \ref{figmselC} and \ref{figmselG}, the model parameters minimizing the error estimates in Theorem \ref{theorem2} also minimizes the true test error. This can be also seen for all other datasets in Table \ref{data} (Appendix B.5). Hence, Theorem \ref{theorem2} provides a practical alternative to resampling techniques for model selection. This is further confirmed from the results in Table \ref{tabBound}. Table \ref{tabBound} shows the average test error over 10 random training/test partitioning of the data in similar proportions as shown in Table \ref{data}. Here, the MU-SVM models selected using Theorem \ref{theorem2} provides similar generalization error compared to the models selected through 5-Fold CV. Further, the proposed model selection strategy using Theorem \ref{theorem2} involves an $O(n+mL)^3$ operation, and provides a computational edge over standard resampling techniques. Table \ref{tabBound}  provides the average time (in seconds) for the MU-SVM model selection using Theorem \ref{theorem2} vs. 5-fold CV for 10 runs over the entire range of parameters. The experiments were run on a desktop with 12 core Intel Xeon @3.5 Ghz and 32 GB RAM. As seen from Table \ref{tabBound}, the bound based model selection is $\sim$ 2-4 times faster than the standard 5-fold resampling technique.

\section{Conclusions} \label{conc}
We introduced a new universum-based formulation for multiclass SVM (MU-SVM). The proposed formulation embodies several useful mathematical properties amenable to: a) an efficient implementation of the MU-SVM formulation using existing multiclass SVM solvers, and b) deriving practical analytic bounds for model selection. We empirically demonstrated the effectiveness of the proposed formulation as well as the bound on real-world datasets. In addition, we also provided insights into the underlying behavior of universum learning and its dependence on the choice of universum samples using the proposed \textit{histogram-of-projections} method.

%

\bibliography{MultiClassUSVM}
\bibliographystyle{icml2018}



\end{document}


\maketitle
\pagestyle{empty}

\section*{Contents}
\begin{itemize}
\item[\textbf{A}] \textbf{Proofs}
	\begin{enumerate}[label=A.\arabic*]
	\item Proof of Proposition 1
	\item Proof of Proposition 2
	\item Derivation of Algorithm 1
	\item Proof of Lemma 1
	\item Proof of Theorem 1
	\item Proof of Proposition 3
	\item Proof of Lemma 2
	\item Proof of Lemma 3
	\item Proof of Theorem 2
	\end{enumerate}
\item[\textbf{B}] \textbf{Additional Results}
	\begin{enumerate}[label=B.\arabic*]
	\item ECOC vs. Direct Approach
	\item SVM vs. MU-SVM using all training classes
	\item Performance comparisons for several Universum types with varying Training set size for GTSRB dataset
	\item Additional Histogram of Projections
	\item Comparison of the error estimates using 5-Fold CV vs. Theorem 2
	\end{enumerate}
\end{itemize}

\appendix

\section{Proofs}
The references cited in this document follows the numbering used in the main paper.

\subsection{Proof of Proposition 1} \label{app::Prop1}
Such a proposition is available for multiclass SVMs (Crammer \& Singer,
2002). Here, we provide a proof for the MU-SVM formulation. Formulation (2) for binary classification becomes,

\begin{flalign}\label{eqUSVMbin}
&\underset{\mathbf{w}_1,\mathbf{w}_2 ,\boldsymbol\xi,\boldsymbol\zeta}{\text{min}}  \quad \frac{1}{2} (\|\mathbf{w}_1 \|_{2}^{2} + \|\mathbf{w}_2 \|_{2}^{2}) + C\sum\limits_{i=1}^n \xi_{i} + C^{*}\sum\limits_{i^\prime =1}^m ( \zeta_{i^\prime 1} + \zeta_{i^\prime 2} )&& \\
&s.t. \quad (\mathbf{w}_{y_{i}}-\mathbf{w}_l)^\top \mathbf{x}_i \geq e_{il} - \xi_{i} ; \quad e_{il} = 1-\delta_{il}, \quad l=1,2 && \nonumber \\
& \quad \quad |(\mathbf{w}_k^\top \mathbf{x}_{i^\prime}^{*}-\underset{l=1,2}{\text{max}}\mathbf{w}_l^\top \mathbf{x}_{i^\prime}^{*})| \leq \Delta + \zeta_{i^\prime k}; \; \zeta_{i^\prime k}, \quad k = 1,2   && \nonumber \\
& \quad \quad i=1\ldots n, \quad i^\prime = 1 \ldots m, \quad \delta_{il} = \left\{
\begin{array}{l l}
    1;\quad  y_i=l \\
    0;\quad  y_i\neq l
\end{array}\right. &&\nonumber
\end{flalign}

The constraints become,\newline
\underline{Training samples ($\forall i = 1 \ldots n$)}  \newline
For any $\mathbf{x}_i \in$ \textbf{class 1} labeled as $y_i = +1$; we have
\begin{flalign}
&(\mathbf{w}_1 - \mathbf{w}_1)^{\top} \mathbf{x}_{i} \geq - \xi_i \quad \Rightarrow \quad \xi_i\geq 0 &&\nonumber \\ 
&(\mathbf{w}_1 - \mathbf{w}_2)^{\top} \mathbf{x}_{i} \geq 1-\xi_i \quad \Rightarrow \quad  y_i(\mathbf{w}_1 - \mathbf{w}_2)^{\top} \mathbf{x}_i \geq 1 - \xi_i &&\nonumber
\end{flalign} 

Similarly, for any $\mathbf{x}_i \in$ \textbf{class 2} labeled as $y_i = -1$; we have,
\begin{flalign}
&(\mathbf{w}_2 - \mathbf{w}_1)^{\top} \mathbf{x}_{i} \geq 1 - \xi_i \quad \Rightarrow \quad y_i(\mathbf{w}_1 - \mathbf{w}_2)^{\top} \mathbf{x}_i \geq 1 - \xi_i  &&\nonumber \\ 
&(\mathbf{w}_2 - \mathbf{w}_2)^{\top} \mathbf{x}_{i} \geq -\xi_i \quad \Rightarrow \quad  \xi_i\geq 0 &&\nonumber
\end{flalign} 

\underline{Universum samples ($\forall i^\prime = 1 \ldots m$)} \newline
For any universum sample $\mathbf{x}_{i^\prime}^{*}$ WLOG we assume $\mathbf{w}_1 \mathbf{x}_{i^\prime}^{*} \geq \mathbf{w}_2 \mathbf{x}_{i^\prime}^{*}$. Then, \newline
\underline{When  k = 1} we have $|\mathbf{w}_1^{\top} \mathbf{x}_{i^\prime}^{*} - \underset{l=1,2}{\text{max}} \mathbf{w}_l^{\top}\mathbf{x}_{i^\prime}^{*}| \leq \Delta+\zeta_{i^\prime k} \quad \Rightarrow \zeta_{i^\prime k} \geq -\Delta$ (true $\because \zeta_{i^\prime k} \geq 0$). \newline

\underline{When  k = 2} we have  $|\mathbf{w}_2^{\top} \mathbf{x}_{i^\prime}^{*} - \underset{l=1,2}{\text{max}} \mathbf{w}_l^{\top}\mathbf{x}_{i^\prime}^{*}| \leq \Delta+\zeta_{i^\prime k} \quad \Rightarrow |\mathbf{w}_2^{\top} \mathbf{x}_{i^\prime}^{*} - \mathbf{w}_1^{\top}\mathbf{x}_{i^\prime}^{*}| \leq \Delta+\zeta_{i^\prime k} \; , \; \zeta_{i^\prime k} \geq 0$.

Hence, eq. \eqref{eqUSVMbin} can be re-written as,
\begin{flalign}\label{eqUSVMbineq1}
&\underset{\mathbf{w}_1,\mathbf{w}_2 ,\boldsymbol\xi,\boldsymbol\zeta}{\text{min}}  \quad \frac{1}{2} (\|\mathbf{w}_1 \|_{2}^{2} + \|\mathbf{w}_2 \|_{2}^{2}) + C\sum\limits_{i=1}^n \xi_{i} + C^{*}\sum\limits_{i^\prime =1}^m \zeta_i^\prime && \\
&s.t. \quad y_i(\mathbf{w}_1-\mathbf{w}_2)^\top \mathbf{x}_i \geq 1 - \xi_{i} ; \quad \xi_{i} \geq 0 , \quad i=1\ldots n&& \nonumber \\
& \quad \quad |(\mathbf{w}_1- \mathbf{w}_2)^\top \mathbf{x}_{i^\prime}^{*}| \leq \Delta + \zeta_{i^\prime}; \quad \zeta_{i^\prime} \geq 0, \quad i^\prime = 1 \ldots m&& \nonumber
\end{flalign}

The solution to the KKT system of \eqref{eqUSVMbineq1} satisfies $\mathbf{w_1} = -\mathbf{w}_2$. Hence replacing $\mathbf{w} = \mathbf{w}_1 - \mathbf{w}_2$ in \eqref{eqUSVMbineq1} still solves \eqref{eqUSVMbin}. This is the U-SVM formulation in (Weston et. al, 2006) with $b=0$. \qed    

\subsection{Proof of Proposition 2} \label{app::Prop2}
The contribution due to the universum samples are same for both (2) and (3). For any universum sample $(\mathbf{x}_{i^\prime}^{*})$ we identify the active constraints and its overall contribution to the objective function through slack variables  i.e. 

\underline{Equation (2)}, the overall contribution of the universum sample $\mathbf{x}_{i^\prime}^{*}$ is,
\begin{flalign}
&C^{*} \sum \limits_{k=1}^L \zeta_{i^\prime k} \quad s.t. \quad |\mathbf{w}_k^{\top} \mathbf{x}_{i^\prime}^{*} - \underset{l=1 \ldots L}{\text{max}}\;\mathbf{w}_{l}^{\top} \mathbf{x}_{i^\prime}^{*}|\leq \Delta+\zeta_{i^\prime k} \quad , \quad \zeta_{i^\prime k} \geq 0, \quad k=1\ldots L &&  \nonumber
\end{flalign}

Case 1: If $k = \underset{l=1\ldots L}{\text{argmax}} \; \mathbf{w}_l^{\top}\mathbf{x}_{i^\prime}^{*}$. The constraint is inactive and $\zeta_{i^\prime k} = 0$.
\newline Case 2: Let $k \neq \underset{l=1\ldots L}{\text{argmax}} \; \mathbf{w}_l^{\top}\mathbf{x}_{i^\prime}^{*}$. Since, $\zeta_{i^\prime k} \geq 0$ the constraint is active if, $-(\mathbf{w}_k^{\top} \mathbf{x}_{i^\prime}^{*} - \underset{l \neq k}{\text{max}}\;\mathbf{w}_{l}^{\top} \mathbf{x}_{i^\prime}^{*})> \Delta$. Then, $\zeta_{i^\prime k}=-[\Delta+(\mathbf{w}_k^{\top} \mathbf{x}_{i^\prime}^{*} - \underset{l \neq k}{\text{max}}\;\mathbf{w}_{l}^{\top} \mathbf{x}_{i^\prime}^{*})]$. 
\newline Hence, keeping only the active constraints the overall contribution of the sample $\mathbf{x}_{i^\prime}^{*}$ is, 
\begin{flalign} \label{eqslack2}
&C^{*} \sum \limits_{k \in \mathcal{K}_{i^\prime}} -[\Delta + \mathbf{w}_k^{\top} \mathbf{x}_{i^\prime}^{*} - \underset{l \neq k}{\text{max}}\;\mathbf{w}_{l}^{\top} \mathbf{x}_{i^\prime}^{*}]
& \quad \text{where,} \quad \mathcal{K}_{i^\prime} = \{ k | -(\mathbf{w}_k^{\top} \mathbf{x}_{i^\prime}^{*} - \underset{l\neq k}{\text{max}}\;\mathbf{w}_{l}^{\top} \mathbf{x}_{i^\prime}^{*}) > \Delta \} && 
\end{flalign} 

\underline{Equation (3)}, Following eq. (3) for the universum sample $\mathbf{x}_{i^\prime}^{*}$ we have L artificial samples as $(\mathbf{x}_{i^\prime}^{*},y_{i^\prime}=1),\ldots , (\mathbf{x}_{i^\prime}^{*},y_{i^\prime}=L) $ stacked at indices $i = n+({i^\prime}-1)L+1 \ldots n+{i^\prime}L$. Hence for $\mathbf{x}_{i^\prime}^{*}$ we have the overall contribution as,
\begin{flalign} 
&C^{*} \sum \limits_{i=n+({i^\prime}-1)L+1}^{n+{i^\prime}L} \xi_{i} \quad \quad \text{s.t.} \; (\mathbf{w}_{y_i} - \mathbf{w}_l)\geq -\Delta (1-\delta_{il}) - \xi_i&&  \nonumber
\end{flalign}   
Now, for $\; i = n+({i^\prime}-1)+k$, we have $\mathbf{x}_i = \mathbf{x}_{i^\prime}^{*}, y_i = k$. The constraints are,
\begin{flalign}
&(\mathbf{w}_k - \mathbf{w}_1)^{\top} \mathbf{x}_{i^\prime}^{*} \geq -\Delta - \xi_i & (\mathbf{w}_k - \mathbf{w}_1)^{\top} \mathbf{x}_{i^\prime}^{*} \geq -\Delta - \xi_i&&\nonumber \\ 
& \quad \quad \vdots & \vdots \quad \quad \quad \quad \quad &&\nonumber\\
&(\mathbf{w}_k - \mathbf{w}_k)^{\top} \mathbf{x}_{i^\prime}^{*} \geq  - \xi_i  \quad \text{(inactive but ensures)} \quad \quad \Rightarrow &  \xi_i \geq 0 \quad \quad \quad \quad  &&\nonumber\\
&\quad \quad  \vdots & \vdots \quad \quad \quad \quad \quad &&\nonumber\\
&(\mathbf{w}_k - \mathbf{w}_L)^{\top} \mathbf{x}_{i^\prime}^{*} \geq -\Delta - \xi_i & (\mathbf{w}_k - \mathbf{w}_L)^{\top} \mathbf{x}_{i^\prime}^{*} \geq -\Delta - \xi_i &&\nonumber
\end{flalign}
This is equivalent to, $-(\mathbf{w}_k^{\top} \mathbf{x}_{i^\prime}^{*} - \underset{l \neq k}{\text{max}} \mathbf{w}_{l}^{\top} \mathbf{x}_{i^\prime}^{*}) \leq \Delta + \xi_i$. Since, $\xi_i \geq 0$ the constraint is active if,$-(\mathbf{w}_k^{\top} \mathbf{x}_{i^\prime}^{*} - \underset{l\neq k}{\text{max}} \mathbf{w}_{l}^{\top} \mathbf{x}_{i^\prime}^{*})>\Delta$, and the contribution becomes, $\xi_i = -[\Delta + \mathbf{w}_k^{\top} \mathbf{x}_{i^\prime}^{*} - \underset{l \neq k}{\text{max}} \mathbf{w}_{l}^{\top} \mathbf{x}_{i^\prime}^{*}]$. Combining all contributions we get, 
\begin{flalign}\label{eqslack4}
&C^{*} \sum \limits_{i=n+({i^\prime}-1)L+1}^{n+{i^\prime}L} \xi_{i} \quad \quad \text{s.t.} \; (\mathbf{w}_{y_i} - \mathbf{w}_l)\geq -\Delta (1-\delta_{il}) - \xi_i&& \nonumber \\
& = C^{*} \sum \limits_{k \in \mathcal{K}_{i^\prime}} -[\Delta + \mathbf{w}_k^{\top} \mathbf{x}_{i^\prime}^{*} - \underset{l \neq k}{\text{max}}\;\mathbf{w}_{l}^{\top} \mathbf{x}_{i^\prime}^{*}]
\quad \text{where,} \quad \mathcal{K}_{i^\prime} = \{ k | -(\mathbf{w}_k^{\top} \mathbf{x}_{i^\prime}^{*} - \underset{l \neq k}{\text{max}}\;\mathbf{w}_{l}^{\top} \mathbf{x}_{i^\prime}^{*}) > \Delta \} && 
\end{flalign}
Comparing \eqref{eqslack2} and \eqref{eqslack4}, the universum sample has similar contribution for both the objective functions in (2) and (4). This is valid for all universum samples. \qed

\subsection{Derivation of Algorithm 1} \label{app::KKTSystem}
In this section we provide the KKT system for (4) and the derivation for the dual form in (5). The proof is available in (Crammer \& Singer, 2002), (Hsu \& Lin, 2002a). We reproduce it for completeness and for better readability of the subsequent proofs. 
The Lagrangian of the MU-SVM formulation is given as, \\
\begin{flalign} \label{eqlag}
\text{Lagrangian, }\mathcal{L} =  \frac{1}{2} \sum \limits_{l}\|\mathbf{w}_l \|_{2}^{2} \;+\; \sum\limits_{i=1}^{n+mL} C_i \ \xi_{i} \; - \; \sum\limits_{il} \eta_{il}[(\mathbf{w}_{y_i} - \mathbf{w}_l)^T \mathbf{x}_i -e_{il} +\xi_i] &&
\end{flalign}
\underline{\textbf{KKT System}}
\begin{flalign} \label{eqkkt}
&\bigtriangledown_{\mathbf{w}_l} \mathcal{L} = 0  \quad \quad \Rightarrow \mathbf{w}_l = \sum \limits_{i} (C_i\delta_{il}- \eta_{il}) \mathbf{x}_i && \\
&\bigtriangledown_{\xi_i} \mathcal{L} = 0  \quad \quad \Rightarrow \sum \limits_{l}\eta_{il} = C_i &&\nonumber 
\end{flalign}
Complimentary Slackness
\begin{flalign} 
&\eta_{il}[(\mathbf{w}_{y_i} - \mathbf{w}_l)^T \mathbf{x}_i -e_{il} +\xi_i] = 0 \quad \forall(i,l) && \nonumber
\end{flalign} 
Constraints,
\begin{flalign} 
&(\mathbf{w}_{y_i} - \mathbf{w}_l)^T \mathbf{x}_i  \geq e_{il} +\xi_i \quad \forall(i,l) && \nonumber\\
&\eta_{il} \geq 0 && \nonumber
\end{flalign}

Finally the dual problem is,
\begin{flalign}\label{eqdual}
\underset{\boldsymbol\eta}{\text{max}} & \quad  - \frac{1}{2} \sum \limits_{i,j} \sum \limits_{l} (C_i\delta_{il} -\eta_{il})(C_j\delta_{jl} -\eta_{jl}) K(\mathbf{x}_i,\mathbf{x}_j) +  \sum\limits_{i,l}\eta_{il}e_{il}  &&\\
s.t. & \quad \sum \limits_{l} \eta_{il} =C_i && \nonumber \\
& \quad \eta_{il} \geq 0 \nonumber&& 
\end{flalign} 
Setting  $\alpha_{il} = C_i\delta_{il} - \eta_{il}$ we get (5). \qed
\newline \newline
 
\subsection{Proof of Lemma 1}
First we prove some interesting properties specific to the MU-SVM solution.

\begin{lemma} $\forall \boldsymbol \alpha_i \in SV_1 = \{i | 0< \alpha_{il} < C_i; y_i=l \}$,
\begin{enumerate}[label=\roman*.]
\item $\sum\limits_{k} \alpha_{ik}[\sum\limits_{j} \alpha_{jk} K(\mathbf{x}_i,\mathbf{x}_j)+e_{ik}] = 0 \; ; \; k = 1 \ldots L$
\item $\forall k \neq y_i$ with $\alpha_{ik}<0 $ \ (strict); \quad $\sum\limits_{j}\alpha_{jk}K(\mathbf{x}_i, \mathbf{x}_j)+e_{ik} = \sum\limits_{j}\alpha_{jy_i}K(\mathbf{x}_i , \mathbf{x}_j)+e_{iy_i}$ i.e. the projection values for the type 1 support vectors for such classes are equal.
\item For any $\gamma_i \in \{ \gamma_i | \sum\limits_{k} \gamma_{ik} = 0 ; \; \gamma_{ik}=0 \; \text{if} \; \alpha_i \in SV_1 \; \text{and} \; \alpha_{ik} = 0 \}$ we have $\sum\limits_k \gamma_{ik}[\sum\limits_j \alpha_{jk} K(\mathbf{x}_i,\mathbf{x}_j)+e_{ik}] = 0$
\end{enumerate}
\end{lemma}
\textbf{Proof} \newline \newline
For simplicity we provide the proof for linear kernel. The same proof applies for non-linear transformations. The proof uses the KKT system for (4).(Appendix \ref{app::KKTSystem})
\begin{enumerate}[label=\roman*.]
\item $\begin{aligned}[t]
&\sum\limits_{k} \eta_{ik}(\mathbf{w}_{y_i}-\mathbf{w}_k)^T \mathbf{x}_i  \quad [\text{From \eqref{eqkkt}}]&&\nonumber \\
=&\sum\limits_{k} \eta_{ik}(\sum\limits_{l}\delta_{il}\mathbf{w}_l)^T\mathbf{x}_i - \sum\limits_{k}\eta_{ik}\mathbf{w}_{k}^T\mathbf{x}_i&&\nonumber \\
=&\sum\limits_{l} C_i\delta_{il}\mathbf{w}_l^T\mathbf{x}_i - \sum\limits_{k}\eta_{ik}\mathbf{w}_{k}^T\mathbf{x}_i \quad = \quad \sum\limits_{k}(C_i\delta_{ik} - \eta_{ik})\mathbf{w}_k^T\mathbf{x}_i&&\nonumber \\
=&\sum\limits_{k}\alpha_{ik} \sum\limits_{j}\alpha_{jk} K(\mathbf{x}_i,\mathbf{x}_j) &&\nonumber 
\end{aligned} $ \newline
From complimentary slackness, if $\alpha_{il} < C_i  \text{with} \; y_i = l \Rightarrow \eta_{il} = (C_i\delta_{il} - \alpha_{il}) > 0  $. This gives, $(\mathbf{w}_{y_i = l} - \mathbf{w}_{k=l})^T \mathbf{x}_i -e_{ik=l} + \xi_i =0 \Rightarrow \xi_i =0$ ( i.e. lies on margin). Now, from complimentary slackness in \eqref{eqkkt},  
\begin{flalign} 
&\sum\limits_{k}\eta_{ik}[(\mathbf{w}_{y_i} -\mathbf{w}_k)^T \mathbf{x}_i -e_{ik}] =0  &&\nonumber \\ 
&\Rightarrow \sum\limits_{k}\alpha_{ik}[\sum_{j}\alpha_{jk}K(\mathbf{x}_i,\mathbf{x}_j)+e_{ik}] = 0 \quad [ \because \eta_{ik}e_{ik}=(C_i\delta_{ik} -\alpha_{ik})e_{ik} = -\alpha_{ik}e_{ik}] \quad  &&\nonumber
\end{flalign}

\item From complimentary slackness \eqref{eqkkt} \\
$\begin{aligned}[t]
& \eta_{ik}[(\mathbf{w}_{y_i} - \mathbf{w}_k)^T\mathbf{x}_i -e_{ik}] = 0 \quad \quad (\forall k \neq y_i \; ; \; \alpha_{ik} < 0 , \because \xi_i = 0) &&\nonumber \\
& \Rightarrow (\mathbf{w}_{y_i} - \mathbf{w}_k)^T\mathbf{x}_i -e_{ik} = 0 \quad \quad (\because \eta_{ik}>0) &&\nonumber\\
&\Rightarrow \mathbf{w}_{y_i}^T \mathbf{x}_i = \mathbf{w}_k^T\mathbf{x}_i+e_{ik}&&\nonumber \\
&\Rightarrow \sum\limits_{j}\alpha_{jy_i}K(\mathbf{x}_i,\mathbf{x}_j)+e_{i y_i} = \sum\limits_{j}\alpha_{j k\neq y_i}K(\mathbf{x}_i,\mathbf{x}_j)+e_{i k \neq y_i}   && \nonumber
\end{aligned}  $ \newline

\item For any such $\gamma_i$, 
\begin{flalign}
&\sum \limits_{k} \gamma_{ik}[\sum \limits_{j} \alpha_{jk} K(\mathbf{x}_i,\mathbf{x}_j)+e_{ik}] &&\nonumber \\ 
=& \gamma_{iy_i}\sum\limits_{j}\alpha_{jy_i}K(\mathbf{x}_i,\mathbf{x}_j) + \sum\limits_{k \neq y_i, \alpha_{ik}<0} \gamma_{ik}[\sum\limits_{j}\alpha_{jk \neq y_i} K(\mathbf{x}_i,\mathbf{x}_j)+e_{ik \neq y_i}]&&\nonumber \\
=&(\gamma_{iy_i}+\sum\limits_{k \neq y_i} \gamma_{iy_i})[\sum\limits_{j} \alpha_{jy_i}K(\mathbf{x}_i,\mathbf{x}_j)] \quad (\text{from ii above and } \because e_{iy_i} = 1-\delta_{iy_i} = 0) &&\nonumber \\
=&0 \quad (\because \sum\limits_{k} \gamma_{ik} = 0 \; \text{by construction}) \quad &&\nonumber
\end{flalign} \qed
\end{enumerate} 
With the above properties for the MU-SVM solution we provide the proof for Lemma 1 following similar lines as in (Vapnik \& Chapelle, 2000). We restate the lemma here for better readability. \newline

\begin{customlemma}{1} 
If in leave-one-out procedure a Type 1 (training) support vector $\mathbf{x}_t \in SV_1\cap \mathcal{T}$ is recognized incorrectly, then we have,
\begin{align}
S_t \ max(\sqrt{2}D,\frac{1}{\sqrt{C}})> 1 \nonumber
\end{align}
where, \begin{flalign}
S_t^2 = &\quad \underset{\boldsymbol\beta}{\text{min}}\quad  \sum \limits_{i,j}(\sum \limits_{l}\beta_{il}\beta_{jl})K(\mathbf{x}_i,\mathbf{x}_j) && \nonumber\\
s.t. & \quad \alpha_{il} - \beta_{il} \leq C_i; \quad \forall \lbrace (i\neq t,l) |\ \alpha_{il} < C_i;\ l = y_i \rbrace &&\nonumber\\
& \quad \alpha_{il} - \beta_{il} \leq 0; \quad \forall\  \lbrace (i\neq t,l) |\ \alpha_{il} > 0;\ l \neq y_i \rbrace &&\nonumber\\
&\quad \beta_{il} =0 ; \quad \forall (i,l) \notin SV_1-\{t\} && \nonumber\\
&\quad \beta_{tl} =\alpha_{tl} ;\quad \forall l && \nonumber\\
&\quad \sum \limits_{l} \beta_{il} =0 &&\nonumber
\end{flalign}
$D$ = Diameter of the smallest hypersphere containing all training samples, and $\mathcal{T}$ = Training set
\end{customlemma}
\textbf{Proof} \newline
The \textit{leave-one-out} formulation for MU-SVM with the $t \in \mathcal{T}$ sample dropped is,
\begin{align}\label{eq10loo}
\underset{\boldsymbol\alpha}{\text{max}} & \quad W(\boldsymbol\alpha)= - \frac{1}{2} \sum \limits_{i,j} \sum \limits_{l} \alpha_{il} \alpha_{jl} K(\mathbf{x}_i,\mathbf{x}_j) -  \sum\limits_{i,l}\alpha_{il}e_{il} \nonumber &&\\
s.t. & \quad \sum \limits_{l} \alpha_{il} =0 &&  \\
& \quad \alpha_{il} \leq C_i \quad  \text{if} \quad l=y_i \quad ; \quad \alpha_{il} \leq 0 \quad  \text{if} \quad l \neq y_i  && \nonumber \\
& \quad \alpha_{tl} = 0 ; \quad  \forall l  \quad \text{(additional constraint)}&& \nonumber
\end{align} 
Then, the \textit{leave-one-out} (l.o.o) error is given as: $ R_{l.o.o} = \frac{1}{n}\sum \limits_{t=1}^{n} \mathbbm{1}[y_t \neq \hat{y}_t] $ where, $\boldsymbol\alpha^t = [\underset{\boldsymbol\alpha_1^t}{\underbrace{\alpha_{11}^t,\ldots,\alpha_{1L}^t}},\ldots,\underset{\boldsymbol\alpha_t^t = \mathbf{0}}{\underbrace{\alpha_{t1}^t = 0,\ldots,\alpha_{tL}^t = 0}},\ldots]$ is the solution for \eqref{eq10loo} and $\hat{y}_t=  \underset{l}{\text{arg max}}\sum \limits_{i}\alpha_{il}^t K(\mathbf{x}_i,\mathbf{x}_t)$ (estimated class label for the $t^{th}$ sample). The overall proof for the bound on the l.o.o error follows three major steps.  \newline

\underline{\textbf{First}}, we construct a feasible solution for (5) using the optimal leave-one-out solution $\boldsymbol\alpha^t$. i.e., construct $\boldsymbol\alpha^t + \boldsymbol\gamma$ as shown below,
\begin{flalign}\label{eq11}
\alpha_{il}^t + \gamma_{il} \leq C_i; \quad  & \forall\ (i,l) \in \lbrace (i,l) | 0 < \alpha_{il}^t < C_i;\ l = y_i \rbrace := A_1^t && \nonumber \\
\alpha_{il}^t + \gamma_{il} \leq 0 ;\quad & \forall\ (i,l) \in \lbrace (i,l) |\ \alpha_{il}^t < 0;\ l \neq y_i \rbrace := A_2^t && \nonumber \\
\gamma_{il} =0;  \quad & \forall (i,l) \notin SV_1^t \quad [SV_1^t = A_1^t \cup A_2^t]&& \nonumber \\
\sum \limits_{l} \gamma_{il} =0 ; \quad \quad &&& 
\end{flalign}
Now,
\begin{flalign} \label{eq12}
I_1 &= W(\boldsymbol\alpha^t + \boldsymbol\gamma)-W(\boldsymbol\alpha^t) && \nonumber\\
&=-\frac{1}{2}\sum \limits_{i,j} \sum \limits_{l}(\alpha_{il}^t + \gamma_{il})(\alpha_{jl}^t + \gamma_{jl})K(\mathbf{x}_i,\mathbf{x}_j)-\sum \limits_{i} \sum \limits_l (\alpha_{il}^t + \gamma_{il})e_{il} + \frac{1}{2} \sum \limits_{i,j} \sum \limits_l \alpha_{il}^t \alpha_{jl}^t K(\mathbf{x}_i,\mathbf{x}_j) + \sum \limits_i \sum \limits_l \alpha_{il}^t e_{il}&& \nonumber \\ 
&= -\frac{1}{2}\sum \limits_{i,j} (\sum \limits_{l} \gamma_{il} \gamma_{jl}) K(\mathbf{x}_i,\mathbf{x}_j)-  \sum \limits_{i,j} (\sum \limits_{l}\gamma_{il} \alpha_{jl}^t)K(\mathbf{x}_i,\mathbf{x}_j)-\sum \limits_{i}\sum \limits_{l}\gamma_{il}e_{il} && \nonumber \\
&= -\frac{1}{2}\sum \limits_{i,j} (\sum \limits_{l} \gamma_{il} \gamma_{jl}) K(\mathbf{x}_i,\mathbf{x}_j)-  \sum \limits_{i,l} \gamma_{il}[\sum \limits_{j} \alpha_{jl}^t K(\mathbf{x}_i,\mathbf{x}_j) +e_{il}] && \nonumber \\
&= -\frac{1}{2}\sum \limits_{i,j} (\sum \limits_{l} \gamma_{il} \gamma_{jl}) K(\mathbf{x}_i,\mathbf{x}_j)-  \sum \limits_{l}\gamma_{tl}[\sum \limits_{j} \alpha_{jl}^t K(\mathbf{x}_j,\mathbf{x}_t)+e_{tl}] \quad (\text{Lemma A.1 (iii)})&& 
\end{flalign}
As a special case we set,
\begin{flalign}\label{eq13}
\boldsymbol \gamma_t&= [\ldots \underset{y_t}{a}, \ldots, \underset{k^{th}}{-a}, \ldots] = a\mathbf{g}_{y_t k} ; \quad  (k = \underset{q \neq y_t}{argmax} \sum\limits_{j} \alpha_{jq}^t K(\mathbf{x}_j,\mathbf{x}_t) \; ; \; \mathbf{g}_{y_t k} = [\ldots \underset{y_t}{1} \ldots \underset{k^{th}}{-1}] )&&  \nonumber
\end{flalign}
Further, we select another $p \in SV_1$ where $\boldsymbol \gamma_{p \neq t} = -a\mathbf{g}_{y_tk}$. Finally, we set, $\boldsymbol \gamma_{i} = 0 \; \forall i \notin \{t,p\}$. For such a case,
\begin{flalign} 
I_1 &= -a^2||\mathbf{x}_t - \mathbf{x}_p||^2 + a[1-(\sum \limits_{j} \alpha_{jy_t}^t K(\mathbf{x}_j,\mathbf{x}_t)-\sum \limits_{j}\alpha_{jk}^t K(\mathbf{x}_j,\mathbf{x}_t))] && \nonumber \\
&\geq -\hat{a}^2 D^2 + \hat{a}[1-(\sum \limits_{j} \alpha_{jy_t}^t K(\mathbf{x}_j,\mathbf{x}_t)-\sum \limits_{j}\alpha_{jk}^t K(\mathbf{x}_j,\mathbf{x}_t))]&&  
\end{flalign}
with, $\hat{a}= \frac{1}{2D^2}[1-(\sum \limits_{j} \alpha_{jy_t}^t K(\mathbf{x}_j,\mathbf{x}_t)-\sum \limits_{j}\alpha_{jk}^t K(\mathbf{x}_j,\mathbf{x}_t))]$ (the value that maximizes the R.H.S in \eqref{eq13}) and $D$ = Diameter of the smallest hypersphere containing all training samples. 
\begin{flalign}
\text{Now, if}; & \quad \hat{a}\leq C \Rightarrow I_1 \geq \frac{1}{4D^2}[1-(\sum \limits_{j} \alpha_{jy_t}^t K(\mathbf{x}_j,\mathbf{x}_t)-\sum \limits_{j}\alpha_{jk}^t K(\mathbf{x}_j,\mathbf{x}_t))]=\frac{1}{2}\hat{a}&&\nonumber \\
\text{else},& \; I_1\geq -C^2D^2+C[1-(\sum \limits_{j} \alpha_{jy_t}^t K(\mathbf{x}_j,\mathbf{x}_t)-\sum \limits_{j}\alpha_{jk}^t K(\mathbf{x}_j,\mathbf{x}_t))]=2CD^2[\hat{a} - \frac{C}{2}] \; \geq 2CD^2\frac{\hat{a}}{2}&&\nonumber
\end{flalign}
If there is an error due to leave one out procedure, then $\underset{q \neq y_t}{max} \sum\limits_{j} \alpha_{jm}^t K(\mathbf{x}_j,\mathbf{x}_t) > \sum \limits_{j} \alpha_{jy_t}^t K(\mathbf{x}_j,\mathbf{x}_t)$. 
\begin{flalign}\label{eq14}
\text{This gives}, \quad I_1 > \frac{1}{2}min(C,\frac{1}{2D^2}) \quad (\text{for l.o.o error})&&
\end{flalign}

\underline{\textbf{Second}}, we construct a feasible solution for the leave-one-out formulation \eqref{eq10loo} using the optimal solution for (5). i.e., construct $\boldsymbol\alpha - \boldsymbol\beta$  as shown below,
\begin{flalign}\label{eq15}
 \alpha_{il} - \beta_{il} \leq C_i; \quad & \forall\ (i,l) \in A_1 -\{t\} ; \quad  A_1 = \lbrace (i,l) |\ 0< \alpha_{il} < C_i;\ l = y_i \rbrace && \nonumber \\
 \alpha_{il} - \beta_{il} \leq 0 ; \quad & \forall\ (i,l) \in A_2 - \{t\}; \quad A_2 = \lbrace (i,l) |\ \alpha_{il} < 0;\ l \neq y_i \rbrace && \nonumber \\
 \sum \limits_{l} \beta{il} =0 ; \quad \quad &&& \nonumber \\
\beta_{il} =0  \quad \quad \quad \quad & \forall (i,l) \notin SV_1-\{t\}&& \\
\boldsymbol \beta_{t} =\boldsymbol \alpha_{t}  \quad \quad \quad \quad &&& \nonumber
\end{flalign} 
with $ \ SV_1 = A_1 \cup A_2\ = \{ i \ | 0< \alpha_{iy_i} < C_i \} $ such that, it is a feasible solution for \eqref{eq10loo}. As before, define	
\begin{flalign}\label{eq16}
I_2 &= W(\boldsymbol\alpha)-W(\boldsymbol\alpha - \boldsymbol\beta) && \nonumber\\
&= -\frac{1}{2} \sum \limits_{i,j} \sum \limits_k \alpha_{il} \alpha_{jl} K(\mathbf{x}_i,\mathbf{x}_j) - \sum \limits_i \sum \limits_l \alpha_{il} e_{il}\quad +\frac{1}{2}\sum \limits_{i,j} \sum \limits_{l}(\alpha_{il} - \beta_{il})(\alpha_{jl} - \beta_{jl})K(\mathbf{x}_i,\mathbf{x}_j)&&\nonumber\\
& \quad +\sum \limits_{i} \sum \limits_l (\alpha_{il} - \beta_{il})e_{il} &&\nonumber\\
&= \frac{1}{2}\sum \limits_{i,j} (\sum \limits_{l} \beta_{il} \beta_{jl}) K(\mathbf{x}_i,\mathbf{x}_j) -\sum\limits_{il}\beta_{il}[\sum \limits_{j}\alpha_{jl}K(\mathbf{x}_j,\mathbf{x_i}) + e_{il}] && \nonumber \\
&= \frac{1}{2}\sum \limits_{i,j} (\sum \limits_{l} \beta_{il} \beta_{jl}) K(\mathbf{x}_i,\mathbf{x}_j) \quad (\text{Lemma A.1 (iii)})&&
\end{flalign} 

\underline{\textbf{Third}}, as the final step define, 
\begin{flalign}\label{eq17}
S_t^2 = &\quad \underset{\boldsymbol\beta}{\text{min}}\quad  \sum \limits_{i,j}(\sum \limits_{l}\beta_{il}\beta_{jl})K(\mathbf{x}_i,\mathbf{x}_j) &&\\
s.t. & \quad \alpha_{il} - \beta_{il} \leq C_i; \quad (i,l) \in A_1 - \{t\}&&\nonumber\\
& \quad \alpha_{il} - \beta_{il} \leq 0; \quad (i,l) \in A_2 - \{t\}&&\nonumber\\
&\quad \beta_{il} =0 ; \quad \forall (i,l) \notin SV_1-\{t\} && \nonumber\\
&\quad \beta_{tl} =\alpha_{tl} ;\quad \forall l && \nonumber\\
&\quad \sum \limits_{l} \beta_{il} =0 &&\nonumber
\end{flalign}
Now, let $\boldsymbol\beta^{\prime}$ be the minimizer for \eqref{eq17}. For such a $\boldsymbol\beta^{\prime}$
\begin{flalign}
I_2 & (=\frac{1}{2}S_t^2)  &&\nonumber \\ 
&\geq I_1  \quad \quad [\because W(\boldsymbol \alpha) \geq W(\boldsymbol \alpha +\boldsymbol \gamma) \;\; \forall \boldsymbol \gamma ; \quad -W(\boldsymbol \alpha - \boldsymbol \beta) \geq -W(\boldsymbol \alpha)\; \; \forall \boldsymbol \beta]&& \nonumber \\
&> \frac{1}{2}min(C,\frac{1}{2D^2}) \quad (\text{from} \eqref{eq14})&& \nonumber 
\end{flalign} \qed
\newline \newline

\subsection{Proof of Theorem 1}
\begin{customthm}{1} 
The leave-one-out error is upper bounded as: 
\begin{flalign} 
R_{l.o.o} \leq & \quad \frac{1}{n} ( |\Psi_1 | + |\Psi_2 |)  && \\
\Psi_2 := & \Big\{ t \in SV_1 \cap \mathcal{T} \; | S_t \ max(\sqrt{2}D,\frac{1}{\sqrt{C}})\geq 1 \Big\} && \nonumber \\
\Psi_1 := & \Big\{ \;t \in  SV_2 \cap \mathcal{T} \Big\} ;\quad |\cdot | := \text{Cardinality of a set} && \nonumber
\end{flalign}
where  $\mathcal{T}:=$ Training Set.
\end{customthm}
\textbf{Proof} \newline
The proof depends on the contribution of a sample to the leave-one-out error, \newline
\underline{\textbf{First}}, for a sample $(\mathbf{x}_t,y_t)$ which is not a support vector, i.e. $t \notin SV$ and $t \in \mathcal{T}$ (Training set); it lies outside margin borders. Dropping such a sample does not change the original solution of (5). Hence, it does not contribute to an error. \newline
\underline{\textbf{Secondly}}, for a sample $(\mathbf{x}_t,y_t) \in SV_1 \cap \mathcal{T}$ contributing to leave-one-out error, Lemma 1 holds i.e. $S_t \ max(\sqrt{2}D,\frac{1}{\sqrt{C}})> 1 $ . \newline
\underline{\textbf{Finally}}, for a sample $(\mathbf{x}_t,y_t)$ with $t \in SV_2 \cap \mathcal{T}$ we add to the leave-one-out error. \qed
\newline \newline

\subsection{Proof of Proposition 3}
\begin{customremark}{2}
If the Type 1 training support vectors i.e. $\{t| t \in SV_1 \cap \mathcal{T}\}$ for SVM and MU-SVM solutions remain same, then we have $S_t^{SVM} \geq S_t^{MU-SVM}$. 
\end{customremark}

\textbf{Proof}
By definition in Lemma 1, \newline \newline
\begin{flalign}
S_t^2 =& \underset{\boldsymbol\beta}{\text{min}}\quad  \sum \limits_{i,j}(\sum \limits_{l}\beta_{il}\beta_{jl})K(\mathbf{x}_i,\mathbf{x}_j)&& \nonumber\\
&s.t. && \nonumber \\
  &\boldsymbol \beta^{MU-SVM} := \left\lbrace
  \begin{array}{cc}
  \alpha_{il} - \beta_{il} \leq C_i; &  (i,l) \in A_1 - \{t\}  \\
  \alpha_{il} - \beta_{il} \leq 0; & (i,l) \in A_2 - \{t\} \\
  \beta_{il} =0 ; & \forall (i,l) \notin SV_1-\{t\} \\
  \beta_{tl} =\alpha_{tl} ;& \forall l \\
  \sum \limits_{l} \beta_{il} =0 & \\
  \end{array}
  \right. \nonumber &&
\end{flalign}

If the Type 1 (training) support vectors for SVM and MU-SVM solutions remain same, we get the same relation as Lemma 1 for C\&S SVM with, 
\begin{flalign}
&\boldsymbol \beta^{SVM} = \{ \beta_{il} \in \boldsymbol \beta^{MU-SVM} | \boldsymbol \beta_i = \boldsymbol \alpha_i \; ; \forall i \in SV_1 \cap \mathcal{U} \} \quad ,\text{where} \; \mathcal{U} = \text{Universum samples}. \nonumber && \\
& \text{i.e.} \; \boldsymbol \beta^{SVM} \subseteq \boldsymbol \beta^{MU-SVM} \Rightarrow S_t(\boldsymbol \beta^{SVM}) \geq S_t(\boldsymbol \beta^{MU-SVM})  && \nonumber
\end{flalign} \qed
\newline \newline

\subsection{Proof of Lemma 2}

\begin{customlemma}{2} 
Under the assumptions (i) and (ii) in Section 3.3 the following equality holds for both Type 1\& 2 training support vectors, i.e. $\mathbf{x}_t \in SV \cap \mathcal{T}$
\begin{align}
S_t^2 =& [\boldsymbol\alpha_t^{\top} \sum\limits_{i \in SV} \sum\limits_{l} \alpha_{il} K(\mathbf{x}_i,\mathbf{x}_t) - \alpha_{ty_t}\mathbf{g}_{y_tk}^{\top} \sum \limits_{i \in SV^t} \sum \limits_{l} \alpha_{il}^t K(\mathbf{x}_i,\mathbf{x}_t)] && \nonumber
\end{align}  with, \quad  $S_t^2 =\{\underset{\boldsymbol\beta}{\text{min}}\ \sum \limits_{i,j}(\sum \limits_{l}\beta_{il}\beta_{jl})K(\mathbf{x}_i,\mathbf{x}_j)|\ \boldsymbol\beta_{t} =\boldsymbol\alpha_{t} ;\ \sum \limits_{l} \beta_{il} =0\ ; (i,j)\in SV_1\}$\quad and \quad  $\mathbf{g}_{y_t k} =[0,\ldots \underset{y_t}{1},\ldots,\underset{k^{th}}{-1},\ldots,0]; \; k = \underset{q \neq y_t}{argmax} \sum\limits_{j} \alpha_{jq}^t K(\mathbf{x}_j,\mathbf{x}_t)$ 
\end{customlemma}
\textbf{Proof} \newline

Under the Assumption (i) we set $\boldsymbol\beta = \boldsymbol\gamma = (\boldsymbol\alpha-\boldsymbol\alpha^{t})$. Then $I_1 = W(\boldsymbol\alpha) - W(\boldsymbol\alpha^t) = I_2$ \\
A similar analysis as in \eqref{eq12} gives,
\begin{flalign} \label{eq18}
I_1 =& -\frac{1}{2}\sum \limits_{(i,j) \in SV_1} (\sum \limits_{l} \gamma_{il} \gamma_{jl}) K(\mathbf{x}_i,\mathbf{x}_j)-  \sum \limits_{l}\alpha_{tl}[\sum \limits_{j \in SV} \alpha_{jl}^t K(\mathbf{x}_j,\mathbf{x}_t)+e_{tl}] && 
\end{flalign}
Note the difference in form compared to \eqref{eq12}. This is because now the analysis applies for both type 1\& 2  support vectors.
Similarly,
\begin{flalign} \label{eq19}
I_2 =& \frac{1}{2}\sum \limits_{i,j} (\sum \limits_{l} \beta_{il} \beta_{jl}) K(\mathbf{x}_i,\mathbf{x}_j) -  \sum \limits_{l}\alpha_{tl}[\sum \limits_{j \in SV} \alpha_{jl} K(\mathbf{x}_j,\mathbf{x}_t)+e_{tl}]  &&
\end{flalign}

Combining, \eqref{eq18} and \eqref{eq19}  
\begin{flalign} \label{eq20}
\sum \limits_{(i,j) \in SV_1} \sum \limits_{l} \beta_{il} \beta_{jl} K(\mathbf{x}_i,\mathbf{x}_j) = \sum \limits_{l}\alpha_{tl}[\sum \limits_{j \in SV} \alpha_{jl} K(\mathbf{x}_j,\mathbf{x}_t)+e_{tl}] -  \sum \limits_{l}\alpha_{tl}[\sum \limits_{j \in SV} \alpha_{jl}^t K(\mathbf{x}_j,\mathbf{x}_t)+e_{tl}] &&
\end{flalign}

Next, let $\boldsymbol\beta^{\prime}$ be the minimizer for \eqref{eq17}. Then,  $(\boldsymbol\alpha-\boldsymbol\beta^{\prime})$ is a feasible solution for \eqref{eq10loo}. Hence,
\begin{flalign} 
&W(\boldsymbol\alpha^t) \geq W(\boldsymbol\alpha-\boldsymbol\beta^{\prime}) &&\nonumber\\
\Rightarrow & W(\boldsymbol\alpha) - W(\boldsymbol\alpha^t) \leq W(\boldsymbol\alpha) - W(\boldsymbol\alpha-\boldsymbol\beta^{\prime}) && \nonumber\\
\Rightarrow & \sum \limits_{i,j}(\sum \limits_{l}\beta_{il}\beta_{jl})K(\mathbf{x}_i,\mathbf{x}_j) \quad \leq \quad S_t^2 && \nonumber
\end{flalign} 
However, from Assumption (i), $\boldsymbol\beta=(\boldsymbol\alpha -\boldsymbol\alpha^t)$ is a feasible solution for \eqref{eq17}. Hence for such a $\boldsymbol\beta$ we have : $ S_t^2 \leq \sum \limits_{i,j}(\sum \limits_{l}\beta_{il}\beta_{jl})K(\mathbf{x}_i,\mathbf{x}_j) $. Combining the above inequality, 
\begin{flalign}  \label{eq21}
S_t^2 = \sum \limits_{i,j}(\sum \limits_{l}\beta_{il}\beta_{jl})K(\mathbf{x}_i,\mathbf{x}_j)&&
\end{flalign}
Further, under Assumption (i) the inequality constraints in \eqref{eq17} are not activated. Hence,  
$S_t^2 =\{\underset{\mathbf{\beta}}{\text{min}}\ \sum \limits_{i,j}(\sum \limits_{l}\beta_{il}\beta_{jl})K(\mathbf{x}_i,\mathbf{x}_j)|\ \boldsymbol\beta_{t} =\boldsymbol\alpha_{t} ;\ \sum \limits_{l} \beta_{il} =0\ ; (i,j)\in SV_1\}$. \newline
Finally combining \eqref{eq20} and \eqref{eq21} we get,
\begin{flalign} \label{eq22}
S_t^2 &= \sum \limits_{l}\alpha_{tl}[\sum \limits_{j \in SV} \alpha_{jl} K(\mathbf{x}_j,\mathbf{x}_t)+\cancel{e_{tl}}] -  \sum \limits_{l}\alpha_{tl}[\sum \limits_{j \in SV} \alpha_{jl}^t K(\mathbf{x}_j,\mathbf{x}_t)+ \cancel{e_{tl}}]
\end{flalign} 
For leave one out error (under Assumption (ii)), 
\begin{flalign}
-\sum \limits_{l}\alpha_{tl}[\sum \limits_{j \in SV} \alpha_{jl}^t K(\mathbf{x}_j,\mathbf{x}_t)] &= \alpha_{ty_t}[\sum \limits_{j \in SV} \alpha_{jk}^t K(\mathbf{x}_j,\mathbf{x}_t) - \sum \limits_{j \in SV} \alpha_{jy_t}^t K(\mathbf{x}_j,\mathbf{x}_t)]  && \nonumber \\
& \geq 0 \quad  \quad \quad (k = \underset{m \neq y_t}{argmax} \sum\limits_{j \in SV} \alpha_{jm}^t K(\mathbf{x}_j,\mathbf{x}_t)) \nonumber &&
\end{flalign}
$\therefore \; S_t^2 \geq \sum \limits_{l}\alpha_{tl}[\sum \limits_{j \in SV} \alpha_{jl} K(\mathbf{x}_j,\mathbf{x}_t)] $ \qed
\newline \newline

\subsection{Proof of Lemma 3}
\begin{customlemma}{3}
The span $S_t^2$ can be efficiently computed as
\begin{flalign} 
S_t^2 &= \left\{
\begin{array}{l l l} 
\boldsymbol\alpha_t^{\top} [(\mathbf{H}^{-1})_{\mathbf{tt}}]^{-1} \boldsymbol\alpha_t \quad & t \in SV_1 \cap \mathcal{T} \\
\boldsymbol\alpha_t^{\top} [K(\mathbf{x}_t,\mathbf{x}_t)\otimes \mathbf{I}_{L}-\mathbf{K}_t^{T}\mathbf{H}^{-1}\mathbf{K}_t] \boldsymbol\alpha_t \quad & t \in SV_2 \cap \mathcal{T}
\end{array} \right. && \nonumber 
\end{flalign}
\begin{flalign}
\text{here,} \quad \mathbf{H} & := \begin{bmatrix}
    \mathbf{K}_{SV_1} \otimes \mathbf{I}_{L}       & \mathbf{A}^{\top} \\
    \mathbf{A}     & \mathbf{0}
\end{bmatrix};  \quad \mathbf{A}:= \mathbf{I}_{|SV_1|} \otimes (\mathbf{1}_L)^{\top} ; \quad \quad  \mathbf{1}_L =[\underset{L\ elements}{\underbrace{1 \ 1 \ldots \ 1}}] \nonumber && \\
\quad (\mathbf{H}^{-1})_{\mathbf{tt}} & := \text{sub-matrix of } \mathbf{H}^{-1} \text{for indices } \quad i = (t-1)L+1  \ldots tL &&\nonumber \\
\quad \mathbf{K}_{SV_1} \quad & := \text{Kernel matrix of Type 1 support vectors}. \quad \text{and} \quad \mathbf{K}_t \quad = [(\mathbf{k}_t^T \otimes \mathbf{1}_L) \; \; \mathbf{0}_{L\times |SV_1|}]^T && \nonumber
\end{flalign}
where, $\mathbf{k}_t = n_{|SV_1| \times 1}$ dim vector where $i^{th}$ element is $K(\mathbf{x}_i,\mathbf{x}_t), \forall \mathbf{x}_i \in SV_1$ ; and $\otimes$ is the Kronecker product.
\end{customlemma}  

\textbf{Proof}

The Span is defined as:
\begin{flalign} 
S_t^2 &= \underset{\mathbf{\boldsymbol\beta}}{\text{min}} \; \sum \limits_{i,j}(\sum \limits_{l} \beta_{il}\beta_{jl})K(\mathbf{x}_i,\mathbf{x}_j) &&  \\ 
& \quad  s.t. \quad \quad \beta_{tl} = \alpha_{tl} \quad ; \quad \forall l = 1, \ldots, L && \nonumber \\
& \quad \quad \quad \quad \sum\limits_{l} \beta_{il} = 0 \quad ; \quad \forall (i,j) \in SV_1 && \nonumber
\end{flalign}

\underline{\textbf{Case}($t \in SV_1$)}
\begin{flalign}
&= \underset{\boldsymbol\beta}{\text{min}} \; \sum \limits_{l} (\alpha_{tl}\alpha_{tl}) K(\mathbf{x}_t,\mathbf{x}_t) + 2 \sum\limits_{i \in SV_1 -\{t\}}\sum \limits_l \alpha_{tl}\beta_{il}K(\mathbf{x}_t,\mathbf{x}_i) + \sum \limits_{(i,j) \in SV_1 - \{t\}}(\sum \limits_{l} \beta_{il}\beta_{jl})K(\mathbf{x}_i,\mathbf{x}_j) && \nonumber \\
& \quad s.t. \quad \underset{\mathbf{A}}{\underbrace{(\mathbf{I}_{|SV_1 - \{t\}|}\otimes \mathbf{1}_L)}} \boldsymbol\beta = \mathbf{0} && \nonumber \\
&=\; \underset{\boldsymbol\beta}{\text{min}}\; \underset{\boldsymbol\mu}{\text{max}}\; \boldsymbol\alpha_t^{\top} [K(\mathbf{x}_t,\mathbf{x}_t) \otimes \mathbf{I}_L] \boldsymbol\alpha_t + 2 \sum\limits_{i\in SV_1 -\{t\}}\sum \limits_l \alpha_{tl}\beta_{il}K(\mathbf{x}_t,\mathbf{x}_i) + \sum \limits_{(i,j) \in SV_1 -\{t\}}(\sum \limits_{l} \beta_{il}\beta_{jl})K(\mathbf{x}_i,\mathbf{x}_j) &&\nonumber\\
& \quad \quad   + 2 \boldsymbol\mu^{\top} \mathbf{A} \boldsymbol\beta + 2\boldsymbol \alpha^T \mathbf{A}_{tt} \boldsymbol \mu \quad \quad \quad \quad  (\boldsymbol\mu := \text{Lagrange Multiplier},\; \because \sum\limits_{l} \alpha_{tl} =0 \Rightarrow \boldsymbol \alpha^T \mathbf{A}_{tt} \boldsymbol \mu =0)&&\nonumber\\
&= \boldsymbol\alpha_t^{\top} [K(\mathbf{x}_t,\mathbf{x}_t) \otimes \mathbf{I}_L] \boldsymbol\alpha_t \; + \; \underset{\boldsymbol\beta}{\text{min}}\; \underset{\boldsymbol\mu}{\text{max}}\; \underset{L(\lambda)}{\underbrace{2\boldsymbol\alpha_t^{\top}(\mathbf{H}_t^{(-t)})^{\top}\boldsymbol\lambda + \boldsymbol\lambda \mathbf{H}^{(-t)}\boldsymbol\lambda}} \quad \quad \quad \quad (\text{with}\quad \boldsymbol\lambda=[\boldsymbol\beta;\boldsymbol\mu]) && \nonumber
\end{flalign}
where,\quad  $\mathbf{I}_{|SV_1 - \{t\}|}:=$ Identity Matrix of size $|SV_1 - \{t\}|$, \newline $\mathbf{A}_{tt} := \; \text{submatrix of} \; \mathbf{A} \text{ for indices} (t-1)L+1,\ldots, tL \;$ \newline $\mathbf{H}^{(-\mathbf{t})} := \; (t-1)L+1,\ldots, tL \; \text{rows/columns of matrix}\; \mathbf{H} $ (in Lemma \ref{lemma3}) removed; and \newline $\mathbf{H}_{\mathbf{t}}^{(-\mathbf{t})}:= (t-1)L+1 ,\ldots, tL \; \text{columns of} \; \mathbf{H}$.  \newline Further, at saddle point :  $\bigtriangledown_{\boldsymbol\lambda}L(\boldsymbol\lambda) = 0 \quad \Rightarrow \boldsymbol\lambda^{*} = - [\mathbf{H}^{(-\mathbf{t})}]^{-1} \mathbf{H}_{\mathbf{t}}^{(-\mathbf{t})} \boldsymbol\alpha_t$. \newline Hence, 
\begin{flalign} 
S_t^2 &= \boldsymbol\alpha_t^{\top} [(K(\mathbf{x}_t,\mathbf{x}_t)\otimes\mathbf{I}_L)-(\mathbf{H}_{\mathbf{t}}^{(-\mathbf{t})})^{\top} (\mathbf{H}^{(-\mathbf{t})})^{-1} \mathbf{H}_{\mathbf{t}}^{(-\mathbf{t})}]\boldsymbol\alpha_t && \nonumber \\
&= \boldsymbol\alpha_t^{\top} (\mathbf{H}^{-1})_{\mathbf{tt}}\boldsymbol\alpha_t && 
\end{flalign} 
where, $(\mathbf{H}^{-1})_{\mathbf{tt}} \quad := \text{sub-matrix of } \;  \mathbf{H}^{-1} \ \text{for index }\;   i\  = (t-1)L+1,\ldots ,tL $.  \newline

\underline{\textbf{Case} ($t \in SV_2$)}
A similar analysis as above gives, 
\begin{flalign} 
S_t^2 &= \alpha_t^{\top} [K(\mathbf{x}_t,\mathbf{x}_t)\otimes \mathbf{I}_{L}-\mathbf{K}_t^{T}\mathbf{H}^{-1}\mathbf{K}_t] \boldsymbol\alpha_t
\end{flalign} 
where, $\mathbf{K}_t \quad = [(\mathbf{k}_t^T \otimes \mathbf{1}_L) \; \; \mathbf{0}_{L\times |SV_1|}]^T \; \text{and} \; \mathbf{k}_t = n_{|SV_1| \times 1}$ dim vector where ith element is $K(\mathbf{x}_i,\mathbf{x}_t), \forall \mathbf{x}_i \in SV_1 $.  \qed
\newline \newline

\subsection{Proof of Theorem 2}
The proof has two steps. 
\begin{itemize}
\item[--] \textit{First}, a sample $(\mathbf{x}_t,y_t)$ which is not a support vector does not contribute to an error.
\item[--] \textit{Secondly}, for a sample $(\mathbf{x}_t,y_t)$ with $t \in SV \cap \mathcal{T}$  Theorem 2 holds. Finally, combining the form of $S_t^2$ in Lemma 3 completes the proof.
\end{itemize} \qed

\section{Additional Results} \label{addResults}

\subsection{ECOC vs. Direct Approach for MU-SVM} \label{app::OVAvsOVO}

This section provides the performance comparisons between two major ECOC based approaches:- \textit{one-vs-all} (OVA) and \textit{one-vs-one} (OVO)  vs. the direct formulation (C \& S based MU-SVM in (2)). For the ECOC based approaches we use standard U-SVM formulation (in Weston et. al 2006) to solve the binary problems. Further,  we use the same datasets and experimental settings as discussed in Section 4. For all the datasets we show the results for Universum types which provided the best performance in Table. 2.

\begin{table*}[h]
\centering
\caption{Mean ($\pm$ standard deviation) test error in \% over 10 runs.} 
\label{tabOVAOVO}
\begin{small}
\begin{sc}
\begin{tabular}{ccccc}  
\hline
\abovespace\belowspace 
\abovespace
Data Set &Method & One Vs All & One Vs All & C\&S (MU-SVM) \\ 
\hline
\abovespace
\multirow{2}{*}{{GTSRB}}&SVM & $7.07 \pm 1.08$ & $7.16 \pm 1.92$ & $7.24 \pm 1.16$ \\ 
\abovespace
&\specialcell{U-SVM (priority-road)} & $6.05\pm 0.61$ & $5.97 \pm 0.63$ & $5.67 \pm 0.32$ \\ \hline
\abovespace
\multirow{2}{*}{{ABCDETC}} & SVM & $28.1 \pm 4.74$ & $29.05 \pm 4.16$ & $27.50 \pm 3.34$ \\ 
\abovespace
&\specialcell{U-SVM (RA)} & $26.05\pm 4.93$ & $26.90 \pm 4.51$ & $22.20 \pm 2.89$ \\ \hline
\abovespace
\multirow{2}{*}{{ISOLET}}&SVM & $3.72 \pm 0.59$ & $3.88 \pm 0.44$ & $3.60 \pm 0.31$ \\ 
\abovespace
&\specialcell{U-SVM (RA)} & $3.56\pm 0.55$ & $3.88 \pm 0.63$ & $2.83 \pm 0.32$ \\ \hline
\end{tabular}
\end{sc}
\end{small}
\end{table*}

As shown above, for the datasets and experimental settings used in this paper, the C\&S based direct formulation (MU-SVM) performs as good as (or better) than the ensemble based methods. 

\subsection{SVM vs MU-SVM using all training classes}\label{app::allclass}

\begin{table*}[h]
\centering
\caption{Performance comparisons between SVM vs. MU-SVM using all training classes}. 
\label{taballclass}
\begin{small}
\begin{sc}
\begin{tabular}{c|cccc}  
\hline
\abovespace\belowspace
\multirow{2}{*}{DataSets}  & \multicolumn{4}{c}{\# Train / Test = 700 / 3500 (100 / 500 per class), \# Universum (m) = 500 } \\
\hline
\abovespace
\multirow{2}{*}{\specialcell{GTSRB}} & \specialcell{MU-SVM \\ (Priority-Road)} & \specialcell{MU-SVM \\(RA)} & \specialcell{MU-SVM \\(Non-Speed)} & - \\ 
\abovespace
SVM = $11.75 \pm 0.77$ & $ 9.77 \pm 0.43$ & $11.29 \pm 0.48$ & $11.82 \pm 0.93$ & - \\ 
\hline
\abovespace\belowspace
&\multicolumn{4}{c}{\# Train / Test = 1500 / 1000 (150 / 100 per class), \# Universum (m) = 300} \\ \hline
\abovespace
\multirow{2}{*}{\specialcell{ABCDETC}} & Upper & Lower & Symbols & RA \\  
\abovespace
 SVM = $42.1 \pm 1.9$ & $41.1 \pm 2.6 $ & $40.2 \pm 3.2 $ & $39.3 \pm 3.2 $  & $38.8 \pm 2.1$  \\
\hline
\end{tabular}
\end{sc}
\end{small}
\end{table*}

\newpage

\subsection{Performance comparisons for several Universum types with varying Training set size for GTSRB dataset}\label{app::varyTrain} 

The experiments follow the same setting as in Table 2. However in this case we vary the number of training samples. The universum set size is fixed to $m = 500$ following  Table 2 i.e. Further, increase in universum samples does not provide significant performance gains. Table \ref{tabGTSRB} provides the mean and std. deviation of the test errors for the SVM and MU-SVM models over 10 random training/test partitioning of the dataset. 

\begin{table*}[h]
\centering
\caption{Mean ($\pm$ standard deviation) of the test errors (in \%) over 10 runs for the GTSRB dataset.} 
\label{tabGTSRB}
\begin{sc}
\begin{tabular}{ccccc}  
\hline
\abovespace\belowspace 
&&\multicolumn{3}{c}{No. of Training samples (per class)}
\\ \cline{3-5}
\abovespace
&Methods  & 300 (100) & 750 (250) & 1500 (500) \\ 
\hline
\abovespace\belowspace
&\specialcell{C\&S SVM} & $7.54 \pm 0.82$ & $4.23 \pm 0.49$ & $3.61 \pm 0.38$ \\ \hline  \\
&\specialcell{\includegraphics[height=0.8cm]{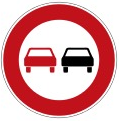}\\(No Passing)} & $6.98 \pm 0.93$ & $4.64 \pm 0.42$ & $3.49 \pm 0.42$ \\ 
\abovespace
&\specialcell{\includegraphics[height=0.8cm]{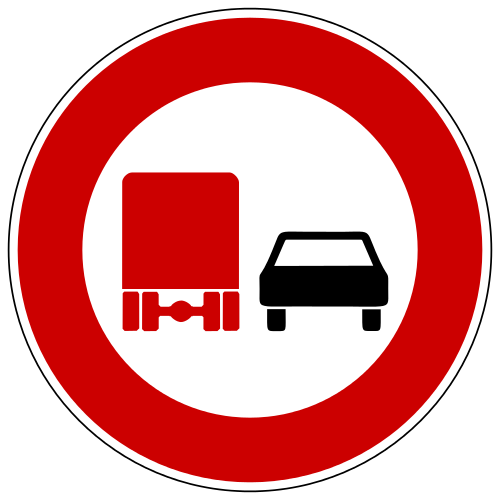}\\(No Passing for Trucks)}  & $6.07\pm 0.68$ & $4.37 \pm 0.9$ & $3.56 \pm 0.41$ \\ 
\abovespace
\multirow{6}{*}{\rotatebox{90}{\specialcell{MU-SVM \\ No. of Universum samsples = 500}}}&\specialcell{\includegraphics[height=0.8cm]{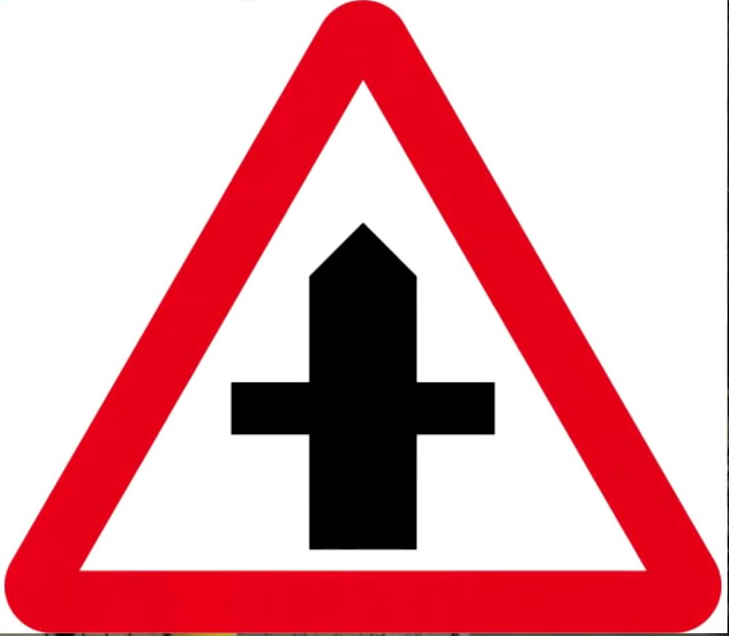}\\(Right of Way)}  & $6.17\pm 0.67$ & $4.03 \pm 0.2$ & $3.12 \pm 0.42$ \\ 
\abovespace
&\specialcell{\includegraphics[height=0.8cm]{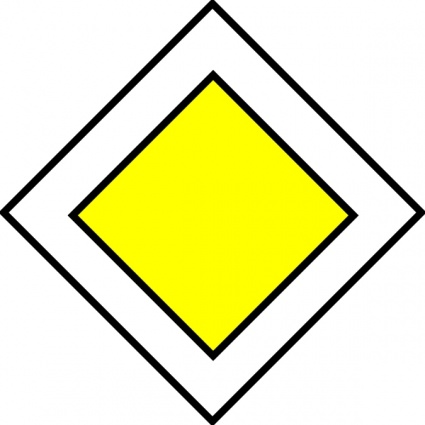}\\(Priority Road)}  & $5.52\pm 0.68$ & $3.52 \pm 0.37$ & $3.15 \pm 0.44$ \\
\abovespace
&\specialcell{\includegraphics[height=0.8cm]{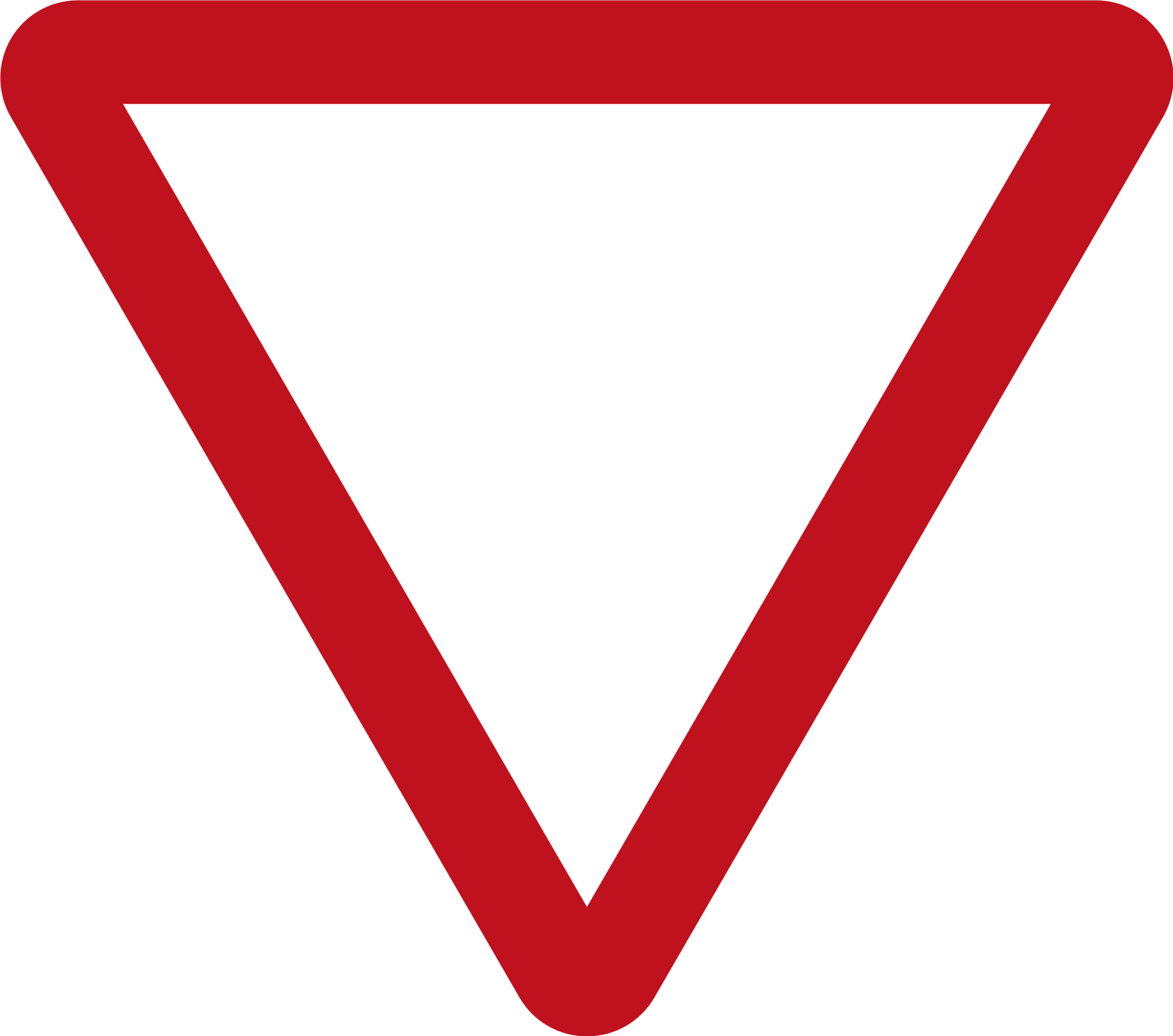}\\(Yield Right of Way)}  & $6.2\pm 0.7$ & $3.83 \pm 0.24$ & $3.11 \pm 0.4$ \\
\abovespace
&\specialcell{\includegraphics[height=0.8cm]{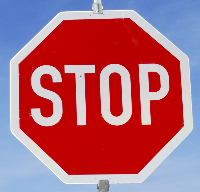}\\(Stop)}  & $6.5\pm 0.66$ & $ 4.24 \pm 0.45$ & $3.21 \pm 0.5$ \\
\abovespace
&\specialcell{\includegraphics[height=0.8cm]{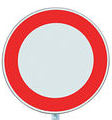}\\(No Vehicles)}  & $6.24\pm 0.39$ & $4.29 \pm 0.33$ & $3.16 \pm 0.24$ \\
\abovespace
&\specialcell{\includegraphics[height=0.8cm]{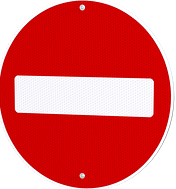}\\(No Entry)}  & $6.17\pm 0.86$ & $3.95 \pm 0.47$ & $3.31 \pm 0.65$ \\
\abovespace
&\specialcell{\includegraphics[height=0.8cm]{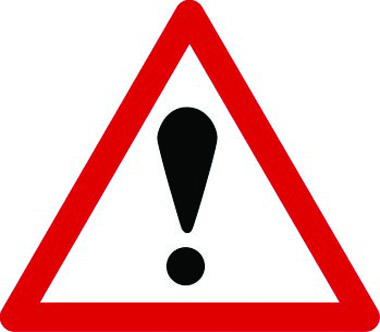}\\(Danger)}  & $6.01\pm 0.74$ & $3.92 \pm 0.55$ & $3.49 \pm 0.62$ \\
\abovespace
&\specialcell{\includegraphics[height=0.8cm]{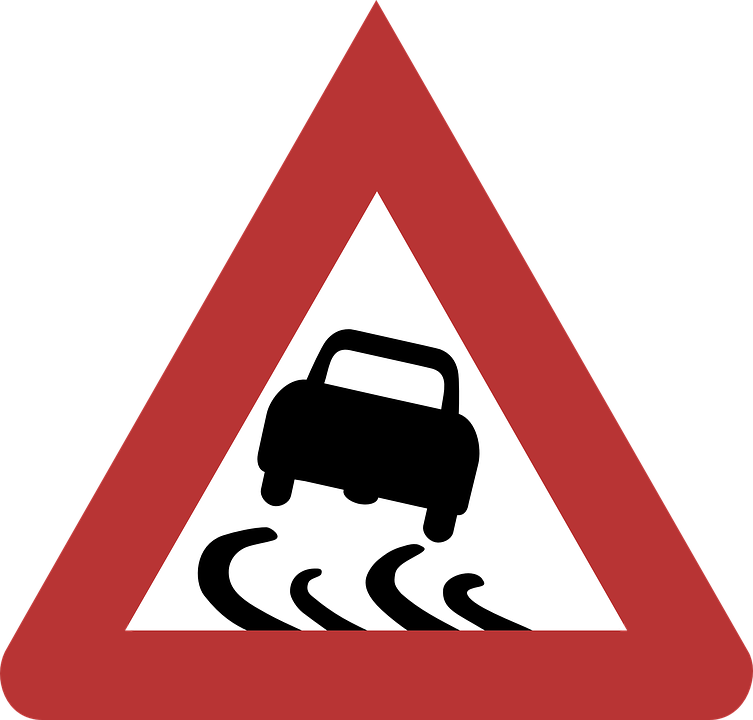}\\(Slippery Road)}  & $6.03\pm 0.64$ & $3.85 \pm 0.28$ & $3.45 \pm 0.62$ \\
&\specialcell{RA} & $6.98 \pm 0.93$ & $4.12 \pm 0.5$ & $3.44 \pm 0.54$ \\ 
&\specialcell{Non Speed} & $7.46 \pm 0.64$ & $4.32 \pm 0.47$ & $3.65\pm 0.4$ \\ \hline
\end{tabular}
\end{sc}
\end{table*}

\begin{landscape}   
 
\begin{figure*}[h]
\centering
\includegraphics[height=2.4cm, width=17cm]{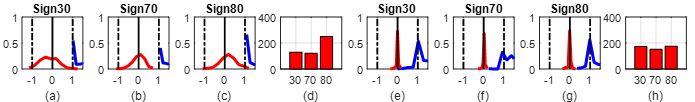}
\caption{Typical histogram of projection of training samples ($n = 750$) (shown in \textcolor{blue}{blue}) and universum samples `\textit{priority-road}' ($m=500$) (shown in \textcolor{red}{red}). SVM decision functions (with $C = 0.1$) for (a) sign `30'. (b) sign `70'.(c) sign `80'. (d) frequency plot of predicted labels for universum samples using SVM model. MU-SVM decision functions (with $C^*/C = 0.5, \Delta = 0.1$) for (e) sign `30'. (f) sign `70'.(g) sign `80'. (h) frequency plot of predicted labels for universum samples using MU-SVM model.} \label{histprior300} 
\end{figure*}

\begin{figure*}[h]
\centering
\includegraphics[height=2.4cm, width=17cm]{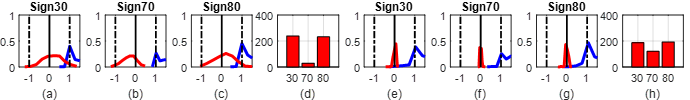}
\caption{Typical histogram of projection of training samples ($n = 1500$) (shown in \textcolor{blue}{blue}) and universum samples `\textit{priority-road}' ($m=500$) (shown in \textcolor{red}{red}). SVM decision functions (with $C = 0.1$) for (a) sign `30'. (b) sign `70'.(c) sign `80'. (d) frequency plot of predicted labels for universum samples using SVM model. MU-SVM decision functions (with $C^*/C = 1, \Delta = 0.05$) for (e) sign `30'. (f) sign `70'.(g) sign `80'. (h) frequency plot of predicted labels for universum samples using MU-SVM model.} \label{histprior500} 
\end{figure*}
 
Table \ref{tabGTSRB} shows that MU-SVM with \textit{priority-road} universum provides the best performance. Further, the performance gains due to MU-SVM reduces with the increase in the number of training samples. For further analysis of this result we use the histogram of projections method. The histogram of projections for the \textit{priority-road} universum with increased training samples $n = 750, 1500$ are provided in Figs. \ref{histprior300} and \ref{histprior500} respectively. As seen from the figures when the number of training samples is large, the estimation problem becomes well-posed and SVM model does not exhibit a huge data-piling effect about the +1 margin borders (compared to Fig. 3). In such cases, application of MU-SVM does not provide a significant improvement over the SVM solution. This is consistent with the results reported in (Cherkassky et al., 2011) for binary U-SVM. This shows that MU-SVM is typically effective for (ill-conditioned) high dimension low sample size settings.

\newpage

\subsection{Additional Histogram of Projections} \label{app::hist}
This section provides the histogram of projections on the modeling results for the ABCDETC and ISOLET datasets. The experimental settings are discussed in Section 4.1.

\subsubsection{ABCDETC Dataset}

\begin{figure*}[h]
\centering
\includegraphics[height=2.8cm, width=22cm]{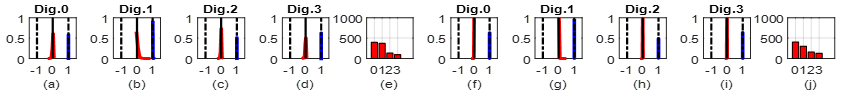}
\caption{Typical histogram of projection of training samples ($n = 600$) (shown in \textcolor{blue}{blue}) and universum samples `\textit{upper case}' letters ($m=1000$) (shown in \textcolor{red}{red}). SVM decision functions (with $C = 1, \gamma = 2^{-7}$) for (a) digit `0'. (b) digit `1'.(c) digit `2'. (d) digit `3'. (e) frequency plot of predicted labels for universum samples using SVM model. MU-SVM decision functions (with $C^*/C = 0.15,\Delta= 0$) for (f) digit `0'. (g) digit `1'.(h) digit `2'. (i) digit `3'.(j) frequency plot of predicted labels for universum samples using MU-SVM model.} \label{histUpper} 
\end{figure*}

\begin{figure*}[h]
\centering
\includegraphics[height=2.8cm, width=22cm]{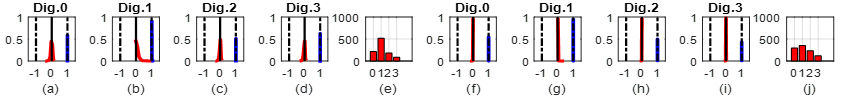}
\caption{Typical histogram of projection of training samples ($n = 600$) (shown in \textcolor{blue}{blue}) and universum samples `\textit{lower case}' letters ($m=1000$) (shown in \textcolor{red}{red}). SVM decision functions (with $C = 1, \gamma = 2^{-7}$) for (a) digit `0'. (b) digit `1'.(c) digit `2'. (d) digit `3'. (e) frequency plot of predicted labels for universum samples using SVM model. MU-SVM decision functions (with $C^*/C = 0.15,\Delta= 0$) for (f) digit `0'. (g) digit `1'.(h) digit `2'. (i) digit `3'.(j) frequency plot of predicted labels for universum samples using MU-SVM model.} \label{histLower} 
\end{figure*}

\begin{figure*}[h]
\centering
\includegraphics[height=2.8cm, width=22cm]{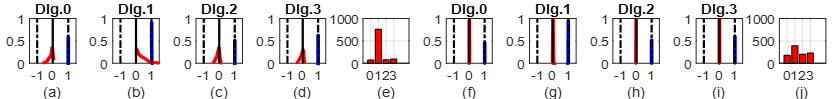}
\caption{Typical histogram of projection of training samples ($n = 600$) (shown in \textcolor{blue}{blue}) and universum samples `\textit{symbols}' ($m=1000$) (shown in \textcolor{red}{red}). SVM decision functions (with $C = 1, \gamma = 2^{-7}$) for (a) digit `0'. (b) digit `1'.(c) digit `2'. (d) digit `3'. (e) frequency plot of predicted labels for universum samples using SVM model. MU-SVM decision functions (with $C^*/C = 0.15,\Delta= 0$) for (f) digit `0'. (g) digit `1'.(h) digit `2'. (i) digit `3'.(j) frequency plot of predicted labels for universum samples using MU-SVM model.} \label{histSymbols} 
\end{figure*}

\begin{figure*}[h]
\centering
\includegraphics[height=2.8cm, width=22cm]{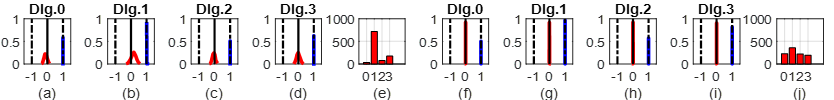}
\caption{Typical histogram of projection of training samples ($n = 600$) (shown in \textcolor{blue}{blue}) and universum samples `\textit{random averaging}' (RA) ($m=1000$) (shown in \textcolor{red}{red}). SVM decision functions (with $C = 1, \gamma = 2^{-7}$) for (a) digit `0'. (b) digit `1'.(c) digit `2'. (d) digit `3'. (e) frequency plot of predicted labels for universum samples using SVM model. MU-SVM decision functions (with $C^*/C = 0.15,\Delta= 0$) for (f) digit `0'. (g) digit `1'.(h) digit `2'. (i) digit `3'.(j) frequency plot of predicted labels for universum samples using MU-SVM model.} \label{histRA} 
\end{figure*}

As seen from Figs \ref{histUpper} - \ref{histRA},
\begin{itemize}
\item \textit{Upper} : the SVM model results in a narrow distribution of the universum samples and in turn provides \textit{near} random prediction on the universum samples. Applying MU-SVM for this case provides no significant change to multiclass SVM solution and hence no additional improvement in generalization (see Table 2).  
\item \textit{Lower} : the SVM model results in a relatively wider distribution of the universum samples (compared to \textit{Upper}). Applying MU-SVM for this case provides some improvement to the multiclass SVM (see Table 2).
\item \textit{Symbol} and \textit{RA} : the SVM model results in a wide distribution of the universum samples. Further, in both the cases the universum samples are mostly predicted as digit `1'. Applying MU-SVM for this case results to a narrow distribution of the universum samples and increases the uncertainity on the universum samples. This results to a significant improvement to the multiclass SVM solution (see Table 2).
\end{itemize}

\subsubsection{ISOLET Dataset}

\begin{figure*}[h]
\centering
\includegraphics[height=2.6cm, width=23cm]{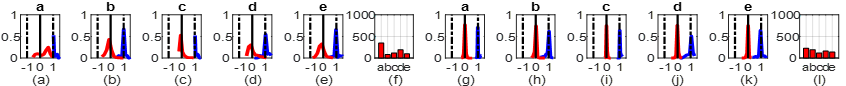}
\caption{Typical histogram of projection of training samples ($n = 500$) (shown in \textcolor{blue}{blue}) and universum samples `\textit{Others}' ($m=1000$) (shown in \textcolor{red}{red}). SVM decision functions (with $C = 1, \gamma = 2^{-7}$) for (a) letter `a'. (b) letter `b'.(c) letter `c'. (d) letter `d'. (e) letter `e'. (f) frequency plot of predicted labels for universum samples using SVM model. MU-SVM decision functions (with $C^*/C = 0.1,\Delta= 0.05$) for (g) letter `a'. (h) letter `b'.(i) letter `c'. (j) letter `d'. (k) letter `e'. (l) frequency plot of predicted labels for universum samples using MU-SVM model.} \label{histOthers} 
\end{figure*}

\begin{figure*}[h]
\centering
\includegraphics[height=2.6cm, width=23cm]{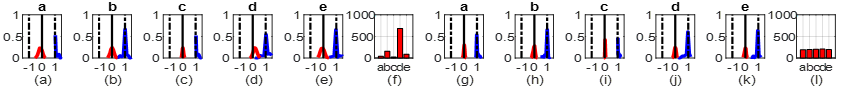} 
\caption{Typical histogram of projection of training samples ($n = 500$) (shown in \textcolor{blue}{blue}) and universum samples `\textit{RA}' ($m=1000$) (shown in \textcolor{red}{red}). SVM decision functions (with $C = 1, \gamma = 2^{-7}$) for (a) letter `a'. (b) letter `b'.(c) letter `c'. (d) letter `d'. (e) letter `e'. (f) frequency plot of predicted labels for universum samples using SVM model. MU-SVM decision functions (with $C^*/C = 0.1,\Delta= 0.1$) for (g) letter `a'. (h) letter `b'.(i) letter `c'. (j) letter `d'. (k) letter `e'. (l) frequency plot of predicted labels for universum samples using MU-SVM model.} \label{histRA_ISOLET} 
\end{figure*}

As seen from Figs \ref{histOthers}-\ref{histRA_ISOLET},
\begin{itemize}
\item \textit{Others} : the SVM model results in a \textit{near} random prediction on the universum samples. Applying MU-SVM for this case reduces the projection of the universum samples but does not result to a significant increase in the uncertaininty of the universum samples, and hence no additional improvement in generalization (see Table 2).  
\item \textit{RA} : the SVM model results in a wide distribution of the universum samples. Further, the universum samples are mostly predicted as letter `d'. Applying MU-SVM for this case results to a narrow distribution of the universum samples and increases the uncertainity on the universum samples. This results to a significant improvement to the multiclass SVM solution (see Table 2).
\end{itemize}

\end{landscape}

\subsection{Comparison of the error estimates using 5-Fold CV vs. Theorem 2}\label{app::errorEst}

This section provides the error estimates curves for the different model parameters for the datasets in Table 1. 

\subsubsection{GTSRB dataset}
\sc with Varying $\;  C^*/C$

\begin{figure*}[h]
\centering
    \begin{subfigure}[t]{4cm}
      \includegraphics[width=4cm]{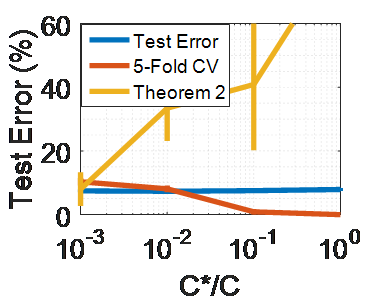}
      \caption{ }
    \end{subfigure}
    \begin{subfigure}[t]{4cm}
      \includegraphics[width=4cm]{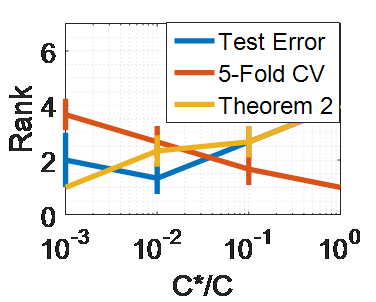}
      \caption{ }
    \end{subfigure}
    \caption{Performance of MU-SVM with \textit{RA} universum for the GTSRB dataset. Here, no. of training samples ($ n = 300 $), no. of universum samples ($m = 1000$) (a) Error estimates for the model parameters $C^*/C = [10^{-3},10^{-2},10^{-1},10^0]$, $C = 1, \;\Delta= 0$. (b) Ranking of the model parameters with the smallest error estimate over each experiments.} \label{GTSRB_SPAN_C_RA}
\end{figure*}

\begin{figure*}[h]
\centering
    \begin{subfigure}[t]{4cm}
      \includegraphics[width=4cm]{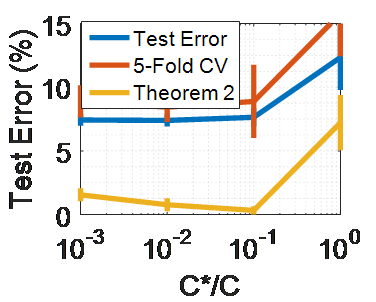}
      \caption{ }
    \end{subfigure}
    \begin{subfigure}[t]{4cm}
      \includegraphics[width=4cm]{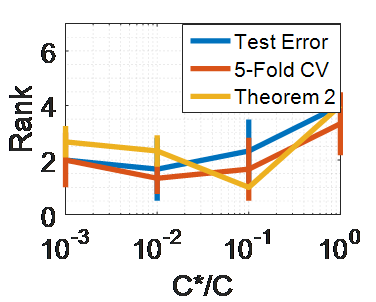}
      \caption{ }
    \end{subfigure}
    \caption{Performance of MU-SVM with \textit{Non-Speed} universum for the GTSRB dataset. Here, no. of training samples ($ n = 300 $), no. of universum samples ($m = 1000$) (a) Error estimates for the model parameters $C^*/C = [10^{-3},10^{-2},10^{-1},10^0]$, $C = 1, \;\Delta= 0$. (b) Ranking of the model parameters with the smallest error estimate over each experiments.} \label{GTSRB_SPAN_C_Others}
\end{figure*}

\sc with Varying $\; \Delta$

\begin{figure*}[h]
\centering
    \begin{subfigure}[t]{4cm}
      \includegraphics[width=4cm]{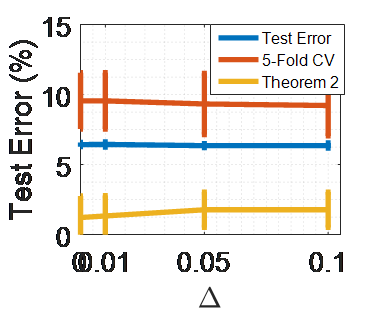}
      \caption{ }
    \end{subfigure}
    \begin{subfigure}[t]{4cm}
      \includegraphics[width=4cm]{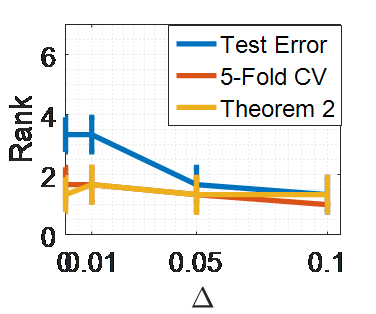}
      \caption{ }
    \end{subfigure}
    \caption{Performance of MU-SVM with \textit{RA} universum for the GTSRB dataset. Here, no. of training samples ($ n = 300 $), no. of universum samples ($m = 1000$) (a) Error estimates for the model parameters $\Delta = [0,0.01,0.05,0.1]$, $C =1, C^*/C = \frac{n}{mL} = 0.1$ (b) Ranking of the model parameters with the smallest error estimate over each experiments.} \label{GTSRB_SPAN_G_RA}
\end{figure*}

\begin{figure*}[h]
\centering
    \begin{subfigure}[t]{4cm}
      \includegraphics[width=4cm]{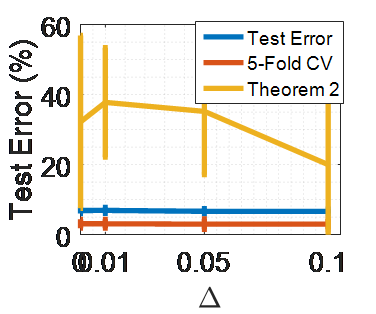}
      \caption{ }
    \end{subfigure}
    \begin{subfigure}[t]{4cm}
      \includegraphics[width=4cm]{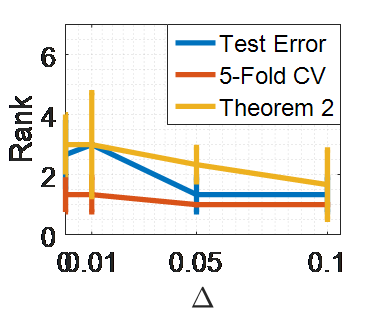}
      \caption{ }
    \end{subfigure}
    \caption{Performance of MU-SVM with \textit{Non-Speed} universum for the GTSRB dataset. Here, no. of training samples ($ n = 300 $), no. of universum samples ($m = 1000$) (a) Error estimates for the model parameters $\Delta = [0,0.01,0.05,0.1]$, $C =1, C^*/C = \frac{n}{mL} = 0.1$ (b) Ranking of the model parameters with the smallest error estimate over each experiments.} \label{GTSRB_SPAN_G_Others}
\end{figure*}
\newpage

\subsubsection{ABCDETC dataset}
\sc with Varying $\;  C^*/C$
\begin{figure*}[h]
\centering
    \begin{subfigure}[t]{4cm}
      \includegraphics[width=4cm]{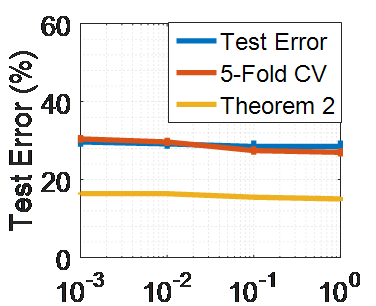}
      \caption{ }
    \end{subfigure}
    \begin{subfigure}[t]{4cm}
      \includegraphics[width=4cm]{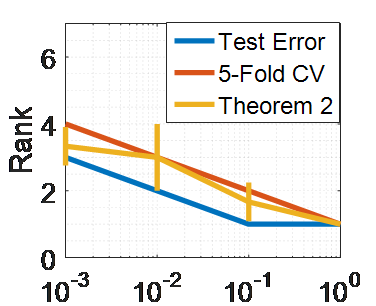}
      \caption{ }
    \end{subfigure}
    \caption{Performance of MU-SVM with \textit{Upper}-case universum for the ABCDETC dataset. Here, no. of training samples ($ n = 600 $), no. of universum samples ($m = 1000$) (a) Error estimates for the model parameters $\Delta = 0$, $C =1, \gamma = 2^{-7}, C^*/C = [10^{-3}, 10^{-2}, 10^{-1}, 10^{0}] $ (b) Ranking of the model parameters with the smallest error estimate over each experiments.} \label{ABCD_Span_C_Upper}
\end{figure*}

\begin{figure*}[h]
\centering
    \begin{subfigure}[t]{4cm}
      \includegraphics[width=4cm]{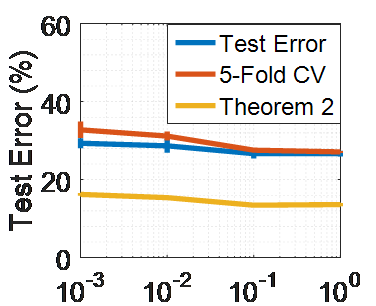}
      \caption{ }
    \end{subfigure}
    \begin{subfigure}[t]{4cm}
      \includegraphics[width=4cm]{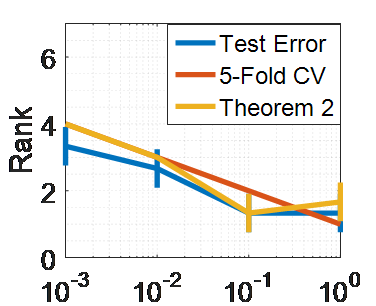}
      \caption{ }
    \end{subfigure}
    \caption{Performance of MU-SVM with \textit{Lower}-case universum for the ABCDETC dataset. Here, no. of training samples ($ n = 600 $), no. of universum samples ($m = 1000$) (a) Error estimates for the model parameters $\Delta = 0$, $C =1, \gamma = 2^{-7}, C^*/C = [10^{-3}, 10^{-2}, 10^{-1}, 10^{0}] $ (b) Ranking of the model parameters with the smallest error estimate over each experiments.} \label{ABCD_Span_C_Lower}
\end{figure*}

\begin{figure*}[h]
\centering
    \begin{subfigure}[t]{4cm}
      \includegraphics[width=4cm]{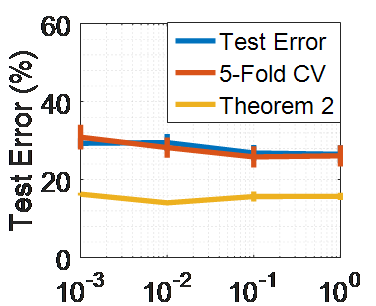}
      \caption{ }
    \end{subfigure}
    \begin{subfigure}[t]{4cm}
      \includegraphics[width=4cm]{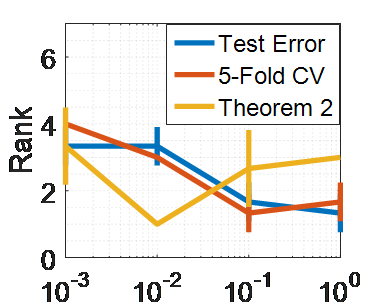}
      \caption{ }
    \end{subfigure}
    \caption{Performance of MU-SVM with \textit{Symbol} universum for the ABCDETC dataset. Here, no. of training samples ($ n = 600 $), no. of universum samples ($m = 1000$) (a) Error estimates for the model parameters $\Delta = 0$, $C =1, \gamma = 2^{-7}, C^*/C = [10^{-3}, 10^{-2}, 10^{-1}, 10^{0}] $ (b) Ranking of the model parameters with the smallest error estimate over each experiments.} \label{ABCD_Span_C_Symbol}
\end{figure*}

\begin{figure*}[h]
\centering
    \begin{subfigure}[t]{4cm}
      \includegraphics[width=4cm]{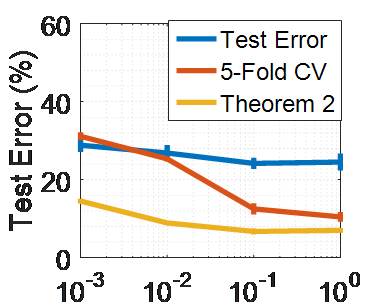}
      \caption{ }
    \end{subfigure}
    \begin{subfigure}[t]{4cm}
      \includegraphics[width=4cm]{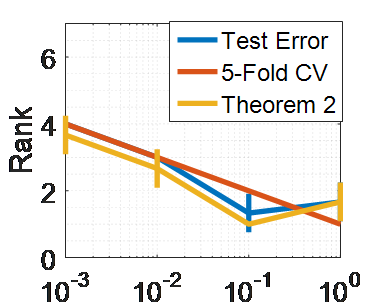}
      \caption{ }
    \end{subfigure}
    \caption{Performance of MU-SVM with \textit{RA} universum for the ABCDETC dataset. Here, no. of training samples ($ n = 600 $), no. of universum samples ($m = 1000$) (a) Error estimates for the model parameters $\Delta = 0$, $C =1, \gamma = 2^{-7}, C^*/C = [10^{-3}, 10^{-2}, 10^{-1}, 10^{0}] $ (b) Ranking of the model parameters with the smallest error estimate over each experiments.} \label{ABCD_Span_C_RA}
\end{figure*}

\newpage

\sc with Varying $\; \Delta$
\begin{figure*}[h]
\centering
    \begin{subfigure}[t]{4cm}
      \includegraphics[width=4cm]{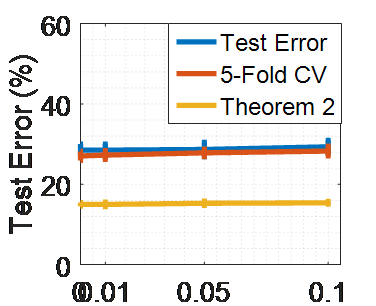}
      \caption{ }
    \end{subfigure}
    \begin{subfigure}[t]{4cm}
      \includegraphics[width=4cm]{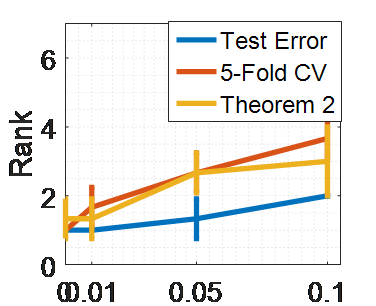}
      \caption{ }
    \end{subfigure}
    \caption{Performance of MU-SVM with \textit{Upper} universum for the ABCDETC dataset. Here, no. of training samples ($ n = 600 $), no. of universum samples ($m = 1000$) (a) Error estimates for the model parameters $\Delta = [0,0.01,0.05,0.1]$, $C =1,\gamma = 2^{-7}, C^*/C = \frac{n}{mL} = 0.15$ (b) Ranking of the model parameters with the smallest error estimate over each experiments.} \label{ABCD_Span_G_Upper}
\end{figure*}

\begin{figure*}[h]
\centering
    \begin{subfigure}[t]{4cm}
      \includegraphics[width=4cm]{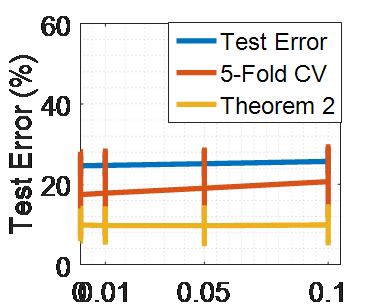}
      \caption{ }
    \end{subfigure}
    \begin{subfigure}[t]{4cm}
      \includegraphics[width=4cm]{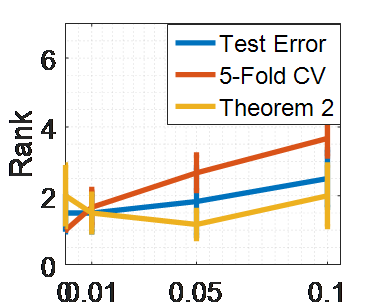}
      \caption{ }
    \end{subfigure}
    \caption{Performance of MU-SVM with \textit{Lower} universum for the ABCDETC dataset. Here, no. of training samples ($ n = 600 $), no. of universum samples ($m = 1000$) (a) Error estimates for the model parameters $\Delta = [0,0.01,0.05,0.1]$, $C =1,\gamma = 2^{-7}, C^*/C = \frac{n}{mL} = 0.15$ (b) Ranking of the model parameters with the smallest error estimate over each experiments.} \label{ABCD_Span_G_Lower}
\end{figure*}

\begin{figure*}[h]
\centering
    \begin{subfigure}[t]{4cm}
      \includegraphics[width=4cm]{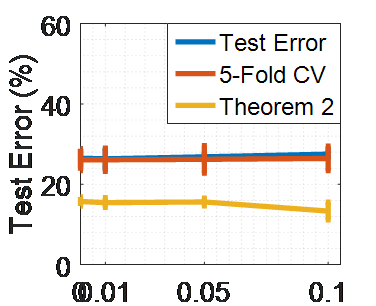}
      \caption{ }
    \end{subfigure}
    \begin{subfigure}[t]{4cm}
      \includegraphics[width=4cm]{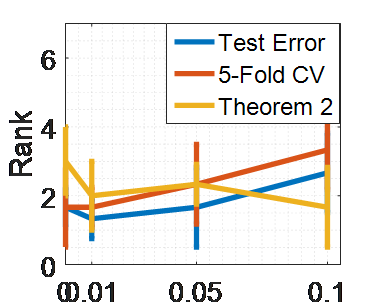}
      \caption{ }
    \end{subfigure}
    \caption{Performance of MU-SVM with \textit{Symbol} universum for the ABCDETC dataset. Here, no. of training samples ($ n = 600 $), no. of universum samples ($m = 1000$) (a) Error estimates for the model parameters $\Delta = [0,0.01,0.05,0.1]$, $C =1,\gamma = 2^{-7}, C^*/C = \frac{n}{mL} = 0.15$ (b) Ranking of the model parameters with the smallest error estimate over each experiments.} \label{ABCD_Span_G_Symbol}
\end{figure*}

\begin{figure*}[!h]
\centering
    \begin{subfigure}[t]{4cm}
      \includegraphics[width=4cm]{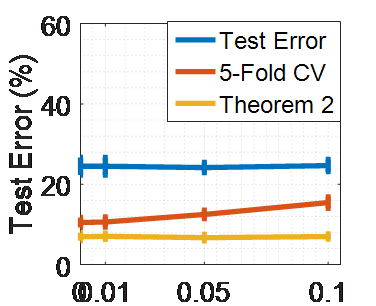}
      \caption{ }
    \end{subfigure}
    \begin{subfigure}[t]{4cm}
      \includegraphics[width=4cm]{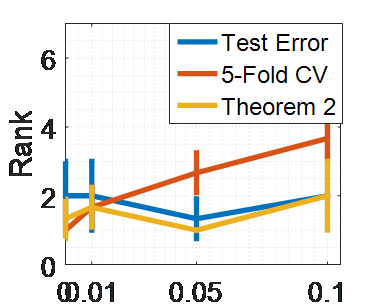}
      \caption{ }
    \end{subfigure}
    \caption{Performance of MU-SVM with \textit{RA} universum for the ABCDETC dataset. Here, no. of training samples ($ n = 600 $), no. of universum samples ($m = 1000$) (a) Error estimates for the model parameters $\Delta = [0,0.01,0.05,0.1]$, $C =1,\gamma = 2^{-7}, C^*/C = \frac{n}{mL} = 0.15$ (b) Ranking of the model parameters with the smallest error estimate over each experiments.} \label{ABCD_Span_G_RA}
\end{figure*}

\newpage
\subsubsection{ISOLET dataset}

\sc with Varying $\; C^*/C$
\begin{figure*}[h]
\centering
    \begin{subfigure}[t]{4cm}
      \includegraphics[width=4cm]{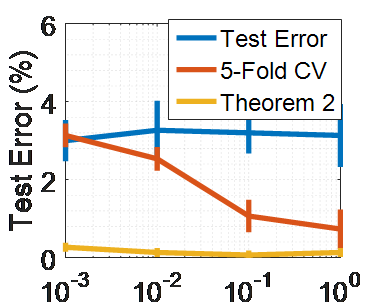}
      \caption{ }
    \end{subfigure}
    \begin{subfigure}[t]{4cm}
      \includegraphics[width=4cm]{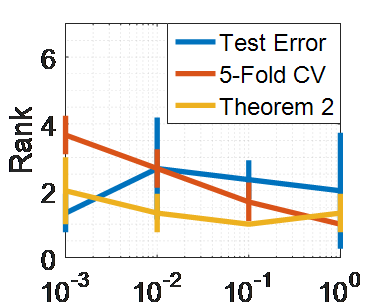}
      \caption{ }
    \end{subfigure}
    \caption{Performance of MU-SVM with \textit{Others} universum for the ISOLET dataset. Here, no. of training samples ($ n = 500 $), no. of universum samples ($m = 1000$) (a) Error estimates for the model parameters $C^*/C = [10^{-3},10^{-2},10^{-1},10^0]$, $C = 1, \;\Delta= 0$ (b) Ranking of the model parameters with the smallest error estimate over each experiments.} \label{ISOLET_Span_C_Others}
\end{figure*}

\begin{figure*}[h]
\centering
    \begin{subfigure}[t]{4cm}
      \includegraphics[width=4cm]{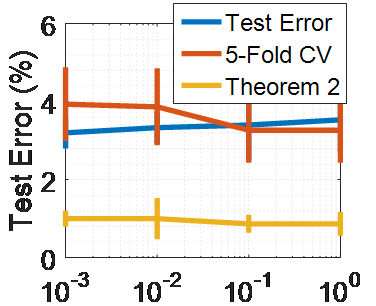}
      \caption{ }
    \end{subfigure}
    \begin{subfigure}[t]{4cm}
      \includegraphics[width=4cm]{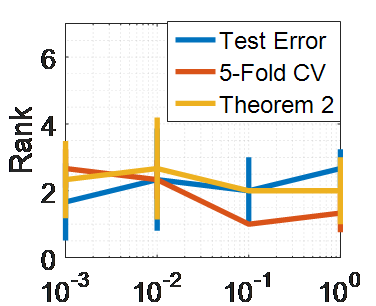}
      \caption{ }
    \end{subfigure}
    \caption{Performance of MU-SVM with \textit{RA} universum for the ISOLET dataset. Here, no. of training samples ($ n = 500 $), no. of universum samples ($m = 1000$) (a) Error estimates for the model parameters $C^*/C = [10^{-3},10^{-2},10^{-1},10^0]$, $C = 1, \;\Delta= 0$ (b) Ranking of the model parameters with the smallest error estimate over each experiments.} \label{ISOLET_Span_C_RA}
\end{figure*}
\newpage

\sc with Varying $\; \Delta$

\begin{figure*}[h]
\centering
    \begin{subfigure}[t]{4cm}
      \includegraphics[width=4cm]{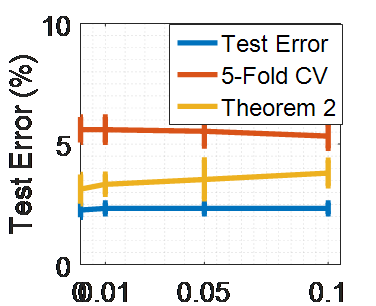}
      \caption{ }
    \end{subfigure}
    \begin{subfigure}[t]{4cm}
      \includegraphics[width=4cm]{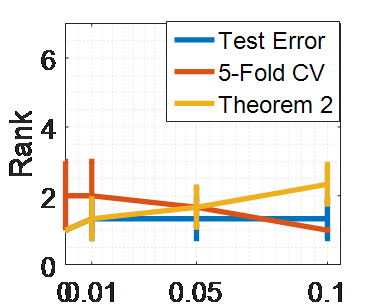}
      \caption{ }
    \end{subfigure}
    \caption{Performance of MU-SVM with \textit{RA} universum for the ABCDETC dataset. Here, no. of training samples ($ n = 600 $), no. of universum samples ($m = 1000$) (a) Error estimates for the model parameters $\Delta = [0,0.01,0.05,0.1]$, $C =1, C^*/C = \frac{n}{mL} = 0.1$ (b) Ranking of the model parameters with the smallest error estimate over each experiments.} \label{ISOLET_Span_G_RA}
\end{figure*}

\begin{figure*}[h]
\centering
    \begin{subfigure}[t]{4cm}
      \includegraphics[width=4cm]{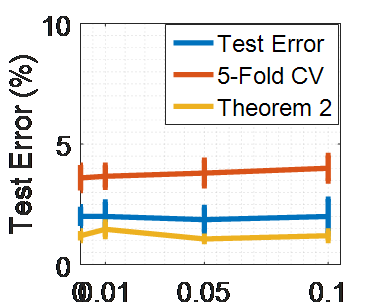}
      \caption{ }
    \end{subfigure}
    \begin{subfigure}[t]{4cm}
      \includegraphics[width=4cm]{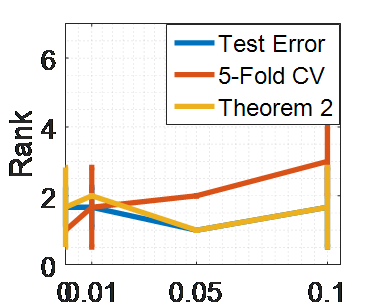}
      \caption{ }
    \end{subfigure}
    \caption{Performance of MU-SVM with \textit{Others} universum for the ABCDETC dataset. Here, no. of training samples ($ n = 600 $), no. of universum samples ($m = 1000$) (a) Error estimates for the model parameters $\Delta = [0,0.01,0.05,0.1]$, $C =1, C^*/C = \frac{n}{mL} = 0.1$ (b) Ranking of the model parameters with the smallest error estimate over each experiments.} \label{ISOLET_Span_G_Others}
\end{figure*}